\documentclass[journal]{IEEEtran}
\usepackage{fix-cm}
\usepackage{array} 
\usepackage{graphicx}%
\usepackage{multirow}%
\usepackage{amsmath,amssymb,amsfonts}%
\usepackage{amsthm}%
\usepackage{mathrsfs}%
\usepackage{xcolor}%
\usepackage{textcomp}%
\usepackage{manyfoot}%
\usepackage{booktabs}%
\usepackage[linesnumbered,ruled,vlined]{algorithm2e}
\usepackage{pdfpages} 
\usepackage{listings}%
\usepackage[utf8]{inputenc}%
\usepackage[T1]{fontenc}%
\usepackage{subcaption}%
\usepackage{amsmath}%

\usepackage[normalem]{ulem}
\usepackage{float}
\usepackage{url}
\usepackage[numbers]{natbib}
\usepackage{hyperref}
\usepackage{caption}

\newcommand\blfootnote[1]{\begingroup\renewcommand\thefootnote{}\footnote{#1}\addtocounter{footnote}{-1}\endgroup}

\begin{document}

\title{Generative Evolutionary Design of Voxel-Based Soft Robots with Provable Optimality}

\author{Junru~Song,~Huan~Xiao,~Yang~Yang,~Guozhen~Li,~Wei~Peng,\\Xiaoya~Zhang,~Tingsong~Jiang,~Weien~Zhou,~Ying~Wen,~Feifei~Wang,~and~Wen~Yao%
\thanks{J.~Song and Y.~Wen are with the School of Computer Science, Shanghai Jiao Tong University, Shanghai 200240, China, and Y.~Wen is also with Shanghai Innovation Institute, Shanghai 200232, China.}%
\thanks{H.~Xiao is with the School of Statistics, Renmin University of China, Beijing 100872, China.}%
\thanks{Y.~Yang, G.~Li, W.~Peng, X.~Zhang, T.~Jiang, W.~Zhou, and W.~Yao are with the Intelligent Game and Decision Laboratory, Beijing 100071, China.}%
\thanks{F.~Wang is with the School of Statistics and the Center for Applied Statistics, Renmin University of China, Beijing 100872, China.}%
\thanks{J.~Song, H.~Xiao, and Y.~Yang contributed equally to this work.}%
\thanks{Corresponding authors: F.~Wang (feifei.wang@ruc.edu.cn) and W.~Yao (wendy0782@126.com).}%
}

\maketitle

\blfootnote{This work has been submitted to the IEEE for possible publication. Copyright may be transferred without notice, after which this version may no longer be accessible.}

\begin{abstract}
Voxel-based soft robots (VSRs) present a promising avenue for developing artificial organisms with lifelike intelligence. However, the vast design spaces and expensive evaluations substantially challenge their design optimization. Here we develop MISCO, a novel evolutionary framework empowered by deep generative models to optimize VSR designs with theoretical guarantees. MISCO integrates an estimation-of-distribution algorithm with a meticulously designed variational autoencoder featuring multi-task learning, position awareness, and inter-voxel signaling. These key components enhance the representational capacity of VSR morphologies and facilitate highly efficient sampling and optimization of morphological distributions. We provide theoretical guarantees for MISCO's asymptotic convergence to globally optimal designs, alongside a favorable convergence rate. Extensive simulated experiments further demonstrate MISCO's exceptional effectiveness in navigating vast design spaces, evolving high-performing VSRs for diverse tasks while flexibly balancing optimization efficiency and morphological diversity. Being validated both empirically and theoretically, MISCO represents a step change towards more scalable and reliable soft robot development. 
\end{abstract}




\section{Introduction}
\label{sec:intro}

\IEEEPARstart{C}{reating} autonomous and adaptive artificial organisms with lifelike intelligence remains a fundamental challenge in robotics \cite{gupta2021embodied}, \cite{blackiston2023biological}, \cite{baines2024robots}. Voxel-based soft robots (VSRs), composed of discrete and deformable units known as voxels, offer a promising solution \cite{hiller2011automatic}, \cite{rossiter2021soft}, \cite{legrand2023reconfigurable}. The modular structure of VSRs enables diverse morphologies and flexible behaviors to emerge, with pronounced resilience and functional versatility. By leveraging various material properties, VSRs further enhance adaptability in dynamic environments and exhibit biomimetic behaviors that resemble those of natural organisms \cite{legrand2023reconfigurable}, \cite{cheney2014unshackling}, \cite{howison2020reality}, \cite{hasanshahi2024design}. Inspired by the well-established philosophy of body-brain co-evolution in embodied cognition \cite{pfeifer2006body}, \cite{pfeifer2014cognition}, \cite{valero2022bio}, VSR development also predominantly adopts a tightly coupled bi-level optimization framework for effective co-adaptation of morphology and control \cite{bhatia2021evolution}, \cite{medvet2021biodiversity}, \cite{song2024morphvae}, \cite{saito2024effective}. Concretely, the outer loop within this framework typically employs population-based evolutionary algorithms (EAs) to iteratively evolve VSR morphologies, while the inner loop dynamically optimizes control policies tailored to each candidate morphology to evaluate fitness for task completion. Driven by advances in both simulators \cite{bhatia2021evolution}, \cite{medvet20202d}, \cite{huang2020dynamic},  \cite{dubied2022sim} and soft materials \cite{legrand2023reconfigurable}, \cite{kriegman2020scalable}, \cite{khodambashi2021miniaturized}, VSR design automation has recently seen notable progress \cite{song2024morphvae}, \cite{saito2024effective}, \cite{liu2023rapidly}, \cite{wang2023preco}, \cite{dong2023leveraging}, \cite{cochevelou2023differentiable}, with an ever-narrowing gap between simulation and physical systems. 

However, the unique advantages of VSRs also present significant challenges for their design automation. \textbf{First}, the non-linear properties of soft materials render VSR design highly abstract and counter-intuitive. This in turn causes domain expertise to fall short and necessitates effective, data-driven evolutionary approaches to fully unlock VSRs' potential \cite{mertan2024investigating}. The intricacy of inter-voxel dynamics further complicates the problem and requires search algorithms to fully characterize these inter-dependencies for effective optimization. \textbf{Second}, the expressive yet combinatorially vast design spaces of VSRs often prevent conventional search heuristics from sufficiently exploring potential design options, leading to premature convergence and sub-optimal solutions \cite{cheney2014unshackling}, \cite{mertan2024investigating}, \cite{stroppa2024optimizing}. \textbf{Third}, the complex fitness landscape of VSRs greatly limits the emergence of useful morphological variability, resulting in either similar evolutionary outcomes \cite{medvet2021biodiversity}, \cite{miras2020environmental}, \cite{pagliuca2022dynamic} or diversified solutions with compromised quality \cite{medvet2021biodiversity}, \cite{pigozzi2023factors}. This quality-diversity dilemma greatly hampers the robustness of robotic systems in volatile environments. \textbf{Last}, the high degrees of freedom inherent in soft materials demand extended control learning and result in notoriously expensive sample evaluation \cite{bhatia2021evolution}, \cite{zhao2024morphological}. It hence becomes a pressing issue to develop more sample-efficient algorithms. One promising direction is to jointly evolve robot designs for multiple tasks simultaneously, with experience transferred across them. However, such \emph{multi-task} design automation remains under-explored. 

Recently, \emph{estimation-of-distribution algorithms} (EDAs) have been shown to resolve some of the above challenges. EDAs are model-based meta-heuristics that explore search spaces by constructing and sampling explicit probabilistic distributions of high-performing solutions \cite{ceberio2024roadmap}. Unlike traditional EAs, EDAs identify promising regions based on discrete, evaluated sample points to promote more comprehensive exploration. The integration of deep generative models (GMs), such as generative adversarial networks (GANs; \cite{goodfellow2020generative}), variational autoencoders (VAEs; \cite{kingma2013auto}) and diffusion models (DMs; \cite{ho2020denoising}), further enhances the representational capacity of EDAs and empowers them to more efficiently navigate high-dimensional, multi-modal landscapes, locating high-performing yet diverse solutions \cite{bhattacharjee2019estimation}, \cite{wittenberg2020dae}, \cite{probst2020harmless}, \cite{wittenberg2022using}. Besides these benefits, the deep generative architectures implement objective-conditional distribution estimation through shared latent space embeddings, enabling simultaneous optimization across multiple design criteria via differentiable sampling from a unified probabilistic model. This formulation establishes deep GM-assisted EDAs as a theoretically grounded framework for sample-efficient soft robot co-design, where the generative model serves as a parametric distribution that adaptively refines the search policy through gradient-based updates. Nevertheless, current studies still face unsolved challenges. For one, they largely rely on task-agnostic neural architectures (\emph{e.g.,} fully-connected layers) \cite{song2024morphvae}, \cite{hu2022modular}, with limited representational capacity of complex interaction dynamics within VSRs. For another, while the value of deep GMs in enhancing diversity has been highlighted, the specific dynamics of the exploration-exploitation trade-off warrant further investigation so as to adaptively adjust evolutionary strategies for various requirements.  

Motivated by these limitations, here we propose \textbf{MISCO}, \underline{m}ult\underline{i}-task \underline{s}oft robot design automation with inter-voxel \underline{co}ordination. MISCO represents, to our best knowledge, the first evolutionary framework for VSR design automation with theoretically underpinned convergence to optimal designs, thus largely ensuring reliability for real-world deployment. The appealing performance of MISCO is primarily driven by a probabilistic generative model meticulously designed for voxel-based soft robots, which we call \textbf{MEC-VAE}, \underline{m}orphological \underline{e}volutionary \underline{c}onditional \underline{V}ariational \underline{A}uto\underline{e}ncoder. MEC-VAE features a modular representation as an inductive bias within its generative process. It explicitly distributes a VSR's overall functionality to individual voxels, taking both general task requirements and each voxel's position into account. These voxel-specific functionalities are subsequently propagated throughout the morphology using Neural Cellular Automata (NCA; \cite{mordvintsev2020growing}) to ensure holistic coordination. This modular architectural design not only achieves a reduced-rank problem formulation, but also significantly promotes the learning of interaction dynamics between voxels. The representational capacity of MEC-VAE is further enhanced by a multi-task learning scheme that maps different tasks to respective VSR distributions with a conditional generative architecture. It hence effectively exploits structural similarities across tasks and allows design experience to be transferred via a shared parameter space. 

To study the exploration-exploitation dynamics within MISCO's evolutionary process, we draw inspiration from an experimental observation previously made in microbial communities, where co-existence of fit and unfit bacteria could contribute to stable diversity \cite{beardmore2011metabolic}. Concretely, we introduce a novel sample-balancing technique, incorporating an adjustable proportion of randomly sampled, sub-optimal robot designs into the population. By modulating the ratio between sub-optimal solutions with top-performing ones, we manage to derive varying evolutionary strategies that strike flexible balances between optimization efficiency and morphological diversity. 

For experimental validation, we choose Evolution Gym (EvoGym; \cite{bhatia2021evolution}) as our simulation platform. We select eight task instances from EvoGym for benchmarking, spanning both locomotion and manipulation with varying difficulty. Based on the adjustable sampling strategy, we propose two variants of MISCO that feature different balances between exploration and exploitation. Specifically, \textbf{MISCO-A} prioritizes the exploitation of \textbf{advantageous} robot designs, whereas \textbf{MISCO-B} emphasizes \textbf{biodiversity} by fitting MEC-VAE exclusively on randomly selected sub-optimal design so as to cover more evolutionary paths. We report that MISCO-A nearly consistently outperforms competitive baselines in terms of optimization efficiency, delivering high-performing solutions with minimal compute. In contrast, MISCO-B achieves substantial improvement in diversity but without sacrificing much of its efficiency. Both variants manifest remarkable inferential capabilities of high-performing VSR distributions, which is attributed mainly to the problem-oriented design of MEC-VAE. We show that this superior property not only fosters more sufficient examination of promising regions within design spaces, but also contributes to more stable evolutionary processes, reflected as stronger directionality towards high performance and fewer unsuccessful attempts. With both empirical and theoretical guarantees, we believe our work non-trivially advances VSR design automation and hopefully inspires future work to study similar evolutionary approaches in broader contexts.

The main contributions of this work are summarized as follows:
\begin{enumerate}
    \item It introduces MISCO, a multi-task evolutionary framework integrating a modular generative architecture, MEC-VAE, which utilizes position awareness and Neural Cellular Automata (NCA) to achieve coordinated inter-voxel signaling across discrete robot lattices.
    \item It proposes a novel sample-balancing strategy inspired by microbial coexistence theory. By adjusting the inclusion of sub-optimal designs, the framework provides a controllable trade-off between rapid exploitation (MISCO-A) and broad morphological biodiversity (MISCO-B).
    \item It establishes the first theoretical guarantee for VSR design with provable optimality. We provide a rigorous proof of asymptotic convergence to global optima (Theorem 1) and demonstrate that the algorithm achieves a linear convergence rate in terms of the Average Convergence Rate (ACR) (Theorem 2).
    \item It validates MISCO’s effectiveness through extensive experiments on the Evolution Gym platform across diverse locomotion and manipulation tasks.
\end{enumerate}

The remainder of this article is organized as follows. Section II reviews the background. Section III discusses technical details of MISCO and its favorable convergence properties. Experimental results and analyses are presented in Section IV. Finally, Section V summarizes our findings and outlines directions for future work. Our code is available at \href{https://github.com/xh621/MISCO}{https://github.com/xh621/MISCO} to ensure reproducibility. 

\begin{figure*}[htbp]
    \centering
    \includegraphics[width=1.0\textwidth]{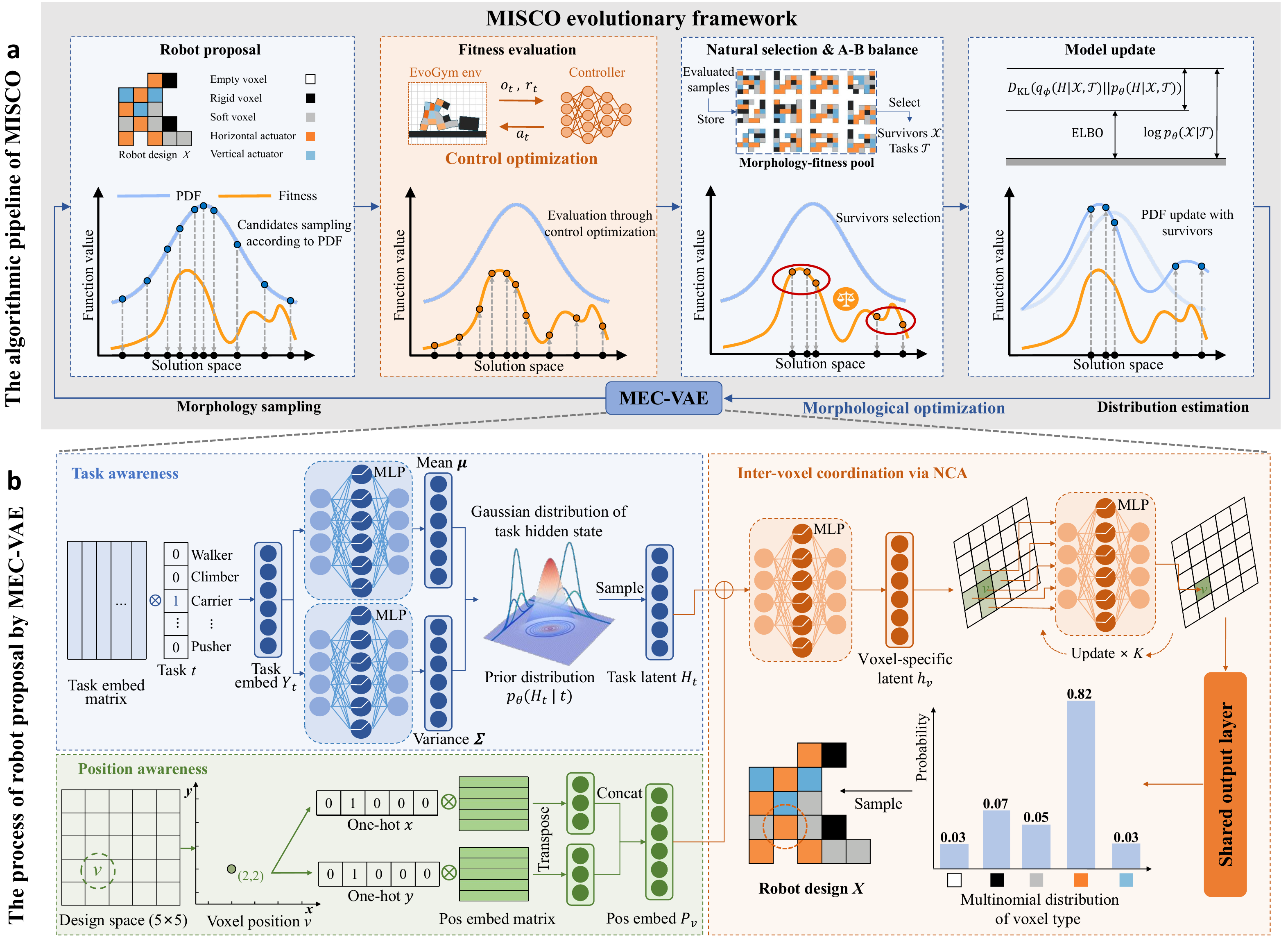}
    \caption{\textbf{Algorithm overview}. \textbf{a,} In each generation, MISCO iterates over four stages: (1) Robot proposal by MEC-VAE; (2) Control optimization and fitness evaluation; (3) Natural selection with an adjustable balance between exploitation and exploration; (4) Model update. \textbf{b,} The process of robot proposal by MEC-VAE.}
    \label{fig:Algorithm Overview}
\end{figure*}

\section{Background}
\subsection{A brief introduction to VSRs}
\label{sec:vsr}
Voxel-based soft robots (VSRs) have garnered significant attention due to their high expressivity, flexibility and biomimetic characteristics. Numerous simulation environments have been developed for VSRs \cite{bhatia2021evolution}, \cite{medvet20202d}, \cite{huang2020dynamic}, \cite{dubied2022sim}. In this work, we focus on the two-dimensional VSRs simulated by Evolution Gym \cite{bhatia2021evolution} for their user-friendly Python interface, fast simulation support and extensive task suite with varying difficulty. As illustrated in the upper left of Fig. \ref{fig:Algorithm Overview}(a), VSRs are represented in a matrix-like layout, where each entry is an integer between 0 and 4, corresponding to one of five voxel types: empty voxel, rigid voxel, soft voxel, horizontal actuator, and vertical actuator. The actuators can actively alter their volumes either horizontally or vertically according to control signals to achieve motion control. Following standard practices in the literature \cite{song2024morphvae}, \cite{saito2024effective}, \cite{liu2023rapidly}, \cite{marzougui2022comparative}, \cite{mertan2023modular}, \cite{songlaser}, we constrain our robot designs to a 5 $\times$ 5 bounding box, which is sufficiently expressive for the emergence of complex and diverse morphological structures while maintaining a tractable search space. 

\subsection{Related work}

Automated VSR design has been explored through a variety of evolutionary strategies \cite{zhang2025soft_robot_design_survey}. Classical approaches such as genetic algorithms \cite{michalewicz2013genetic} and Bayesian optimization \cite{kandasamy2018parallelised} iteratively refine morphological candidates but often converge prematurely in the combinatorially vast and deceptive fitness landscapes of VSRs \cite{mertan2024investigating}, \cite{stroppa2024optimizing}. Generative encodings, exemplified by CPPN-NEAT \cite{stanley2007compositional}, \cite{cheney2014unshackling}, alleviate this issue by promoting phenotypic regularity, yet their indirect representations limit the transparency and controllability of the design process. More recent efforts have explored differentiable \cite{cochevelou2023differentiable}, \cite{strgar2024evolution}, \cite{kobayashi2024topology} and curriculum-based \cite{wang2023curriculum}, \cite{wang2023preco} strategies to improve co-optimization efficiency. Despite their merits, these methods typically operate on fixed morphological parameterizations and do not maintain an explicit probability model over the design space, limiting their ability to systematically reason about the distribution of high-performing designs.

Estimation of Distribution Algorithms (EDAs) \cite{hauschild2011introduction}, \cite{ceberio2024roadmap}, \cite{larranaga2024eda_survey} address this limitation by constructing and sampling explicit probabilistic models of promising solutions. Their recent integration with deep generative models has enabled more expressive distribution estimation over structured search spaces \cite{bhattacharjee2019estimation,wittenberg2022using,zhang2026opt}. RoboGAN \cite{hu2022modular} applied generative adversarial networks to EDA-based modular robot design, training a GAN to approximate the distribution of high-fitness morphologies via adversarial learning. MorphVAE \cite{song2024morphvae} adopted variational autoencoders \cite{kingma2013auto} for VSR design, demonstrating that continuous latent manifolds smooth the optimization landscape and enable multi-task knowledge sharing. However, these methods rely on task-agnostic, fully-connected architectures that treat the morphology as a flat vector, overlooking the spatial structure and local interaction dynamics intrinsic to VSRs. Since a voxel's optimal material type depends critically on its neighbors' configurations, this architectural limitation constrains the fidelity with which these models can represent high-performing morphological distributions.

Neural Cellular Automata (NCA; \cite{mordvintsev2020growing}) offer a complementary perspective by modeling decentralized developmental processes through iterative local update rules parameterized by neural networks. Horibe et al.\ \cite{horibe2021regenerating} demonstrated that NCA-based developmental encodings can endow soft robots with regenerative capabilities, enabling damaged morphologies to self-repair. Nevertheless, existing applications of NCA in soft robotics remain confined to standalone developmental schemes and have not been integrated into distribution-based evolutionary frameworks, leaving their potential as structural inductive biases within generative probabilistic models largely untapped.

Two important open problems further constrain the maturity of deep GM-assisted EDAs. First, convergence guarantees for EDAs have been established only for simple probabilistic models: Zhang and Sun \cite{zhang2004convergence} proved global convergence under smoothed updating for classical EDAs, Friedrich and K\"otzing \cite{friedrich2020eda} refined the analysis for univariate cases, and Chen and He \cite{chen2021average} introduced the Average Convergence Rate (ACR) as an efficiency metric. These results, however, do not extend to deep generative models, leaving a gap between empirical success and theoretical reliability. Second, the quality-diversity (QD) dilemma \cite{pugh2016qd}---the tension between optimization efficiency and solution diversity---remains difficult to manage in practice. Morphological diversity in evolved VSR populations tends to collapse under selection pressure \cite{medvet2021biodiversity}, \cite{pigozzi2023factors}, and existing countermeasures either require predefined behavioral descriptors that are difficult to specify for complex morphologies, or sacrifice optimization performance for coverage. While evidence from microbial ecology suggests that coexistence of fit and unfit strains can stabilize biodiversity through metabolic trade-offs \cite{beardmore2011metabolic}, this biological insight has yet to be systematically leveraged in evolutionary robot design.

\section{Methods}

\subsection{Overview of MISCO}
In this work, we propose \textbf{MISCO}, \underline{m}ult\underline{i}-task \underline{s}oft robot design automation with inter-voxel \underline{co}ordination. As demonstrated in Fig. \ref{fig:Algorithm Overview}(a), MISCO employs a bi-level optimization scheme and relies on \textbf{MEC-VAE}, the \underline{m}orphological \underline{e}volutionary \underline{c}onditional \underline{v}ariational \underline{a}uto\underline{e}ncoder, for the modeling of morphological distributions. Specifically, the inner loop of MISCO involves control optimization of candidate designs to evaluate their fitness, while MEC-VAE serves as the foundation of the outer loop by fitting high-performing designs and proposing offspring solutions in each generation. Each iteration of MISCO can be decomposed into the following four stages and outlined in Alg. \ref{Algorithmic pseudo-code}. Each of the stages is detailed in the following subsection \ref{sec:Robot proposal} to subsection
 \ref{sec:model update}. 

\begin{enumerate}
    \item \textbf{Robot proposal - }MEC-VAE generates a population of robot morphologies for each task through task awareness, position awareness and inter-voxel coordination. 
    \item \textbf{Fitness evaluation - }Each morphology is placed in its corresponding simulation environment to interact, and an optimized control policy is obtained via reinforcement learning. Each morphology and its policy constitute an intelligent agent, with its cummulative reward serving as the fitness evaluation. The evaluated morphologies and their fitness scores are stored in a sample pool.
    \item \textbf{Natural selection - }A set of high-performing robot morphologies is selected from the sample pool, with an adjustable balance between exploitation and exploration. 
    \item \textbf{Model update - }The parameters of MEC-VAE are then updated based on the selected samples to better approximate the distribution of high-performing morphologies. The algorithm then returns to the first stage and iterates until the termination condition (\emph{e.g.,} a maximal number of robot evaluations) is met. 
\end{enumerate}

To our knowledge, MISCO establishes the first theoretical guarantee for the convergence of deep GM-assisted EDAs. We show that under mild conditions, MISCO will almost surely converge to optimal distributions, which assign non-zero probabilities only to globally optimal robot designs. We further examine the average convergence rate of MISCO and show that it enjoys \emph{linear} convergence. Intuitively, this indicates that the discrepancy between the current population and optimality would decrease exponentially with time. Therefore, MISCO not only exhibits asymptotic convergence, but also requires less compute to find optimal (or near-optimal) solutions, ensuring sample efficiency in practical application. Please see subsection \ref{sec:Theoretical guarantees of convergence} for a formal analysis.

\subsection{Robot proposal}
\label{sec:Robot proposal}
The performance of an EDA largely relies on the statistical model it uses to approximate high-performing solution distributions. While previous studies have yielded promising results by leveraging deep generative models, we note that their overly simple architectures do not suffice for effective VSR design. To this end, we introduce MEC-VAE, a Variational Autoencoder specifically designed for VSR modeling. MEC-VAE serves as the core of our evolutionary framework and consists of two major parts: an encoder for posterior approximation and a decoder for robot morphology proposal. 

The encoder approximates the posterior distribution of latent variables given a specific robot morphology and task type, facilitating variational inference. More details about the encoder and its role in MEC-VAE’s model update process can be find in subsection \ref{sec:model update}. Here we will focus on the decoder module.

The decoder simulates the generative process by incorporating task-specific inductive biases through three integrated components: task awareness, position awareness, and inter-voxel coordination. In the following, we describe these components in detail, highlighting how they gradually enable well-coordinated voxel arrangements. These components are progressively applied in the generative process as illustrated in Fig. \ref{fig:Algorithm Overview}(b): For each voxel position, MEC-VAE first applies learnable embedding layers to capture task-related and position-related information. These embeddings are then integrated through a feedforward neural network to derive voxel-specific latent $h_v$. Next, inter-voxel coordination takes place between neighboring voxels through Neural Cellular Automata (NCA; \cite{mordvintsev2020growing}), where the $h_v$'s are iteratively updated over $K$ steps. The resulting latents, \( H_v \)'s, are then passed through a shared output layer to generate a multinomial distribution over possible material types. Sampling from these distributions determines the material types across the entire body plan, leading to a complete design proposal. 

\subsubsection{Task awareness}
As robot morphology is highly dependent on the specific task it is designed for, the generative process must be task-aware in the first place. To achieve this, each task $t$ is assigned a learnable embedding vector $Y_t$ that captures its unique characteristics. These embedding vectors are then used to derive the prior distribution of task-aware latent variables $H_t$. By mapping the deterministic task embeddings into latent Gaussian distributions, we allow each task to sample from a host of various morphological prototypes, thus achieving more realistic and diversified morphogenesis modeling. The mean vector and covariance matrix of $H_t$ are both computed using multi-layer perceptrons (MLPs) that take $Y_t$ as input, and denoted as $\mu_\theta(Y_t)$ and $\Sigma_\theta(Y_t)$, respectively. The above calculation is concisely expressed in equation (\ref{eq:task awareness})
\begin{equation}
\label{eq:task awareness}
    Y_t = \text{Task\_Embed\_Layer}(t),\ 
    H_t \sim \mathcal{N}(\mu_{\theta}(Y_t), \Sigma_{\theta}(Y_t))
\end{equation}

\subsubsection{Position awareness}

Different modules within a robot, analogous to different body parts of a biological organism, should specialize in distinct functions depending on their positions in the body. We represent the positional information of different modules with learnable embedding vectors. For each module $v$, its one-hot encoded coordinates along the $x$- and $y$-axes of the material matrix, denoted as $c_x$ and $c_y$ respectively, are processed by an embedding layer to obtain positional representations and then concatenated to yield $P_v$. 
\begin{equation}
\begin{split}
    P_v = \text{Concat}(\text{Position\_Embed\_Layer}(c_x),\\ \text{Position\_Embed\_Layer}(c_y))
\end{split}
\end{equation}

\subsubsection{Inter-voxel coordination}

Modules within a robot must interact and coordinate seamlessly to achieve a unified objective. To this end, we employ Neural Cellular Automata (NCA; \cite{mordvintsev2020growing}) as a message-passing scheme to enable efficient information exchange between neighboring voxels. The positional embeddings $P_v$ are first concatenated with the task-aware latent vectors $H_t$ and then processed with an MLP denoted as $\phi_\theta$. This results in a unique latent vector $h_v$ for each voxel that integrates both the overall task requirements and the voxel's specific functionality. Information is then transmitted between adjacent voxels via NCA, strengthening inter-voxel coordination and yielding the final latent vectors $H_v$. We note that this inter-voxel message passing scheme greatly resembles bioelectric signaling networks in developmental biology that enable coordinated gene expression across cells for complex anatomical outcomes. 

Unlike traditional cellular automata, where the state of each cell is updated based on predefined rules, NCA utilizes rules parameterized by neural networks that are trainable via gradient-based optimization. This approach allows NCA to model complex behaviors and dynamics that are difficult to capture with hand-crafted rules. In our framework, each module’s updating rule is represented as a \emph{shared} MLP that takes its own state and the states of neighboring modules as input to compute the next state (as shown in Fig. \ref{fig:Algorithm Overview}(b) and equation (\ref{eq:NCA})). In this way, we effectively exploit the homogeneity of local communication patterns to minimize computational cost, while holistically coordinating different modules to serve unified objectives within multiple rounds of communication. This iterative message-passing process (as shown in equation (\ref{eq:NCA detail})) continues for a predefined number of $K$ steps until the latent vectors of different voxels, $h_v$, evolve into a steady state denoted as $H_v$. 
\begin{equation}
\label{eq:NCA}
    h_v = \phi_\theta(\text{Concat}(H_t, P_v)), \quad H_v = \text{NCA}(h_v, h_{\text{neighbor}})
\end{equation}

The detailed computation of NCA is as follows: 
\begin{equation}
\label{eq:NCA detail}
\begin{split}
h_v^{k+1} = \text{Concat} \left( \text{MLP}(h_v^k, \ h_{neighbor}^k), \ h_v^k \right),\\ \ k = 0, 1, \dots, K-1 
\end{split}
\end{equation}
where $h_v^0=h_v$, $H_v=h_v^K$, $h_{\text{neighbor}}$ denotes concatenated states of neighboring voxels (\emph{i.e.,} up, down, left and right) with absent neighbors replaced with zero vectors. The concatenation of $h_v^k$ in each time step serves as skip connection that promotes easier gradient flows. 

\subsubsection{Voxel sampling}

The final latent representations $H_v$ are fed into the output layer, which is also an MLP, to determine the material type of each voxel (equation (\ref{eq:sampling})). Since $H_v$ already contains heterogeneous functionality representation that is specific to each voxel $v$, the output layer is shared across all voxel positions to avoid parameter proliferation. Specifically, the output layer, denoted as $F_\theta$, takes $H_v$ as input and outputs the parameters of a multinomial distribution $(p_{v,1}, p_{v,2}, p_{v,3}, p_{v,4}, p_{v,5})$. Here $p_{v,1}$ through $p_{v,5}$ correspond to the five voxel types introduced in subsection \ref{sec:vsr}, with their probabilities summing to one. By sampling each voxel and combining the results of all $5\times5=25$ positions, a complete robot design $X$ is eventually assembled. 
\begin{equation}
\label{eq:sampling}
\begin{split}
p_{v,1}, p_{v,2}, p_{v,3}, p_{v,4}, p_{v,5} = F_\theta(H_v), \\ \quad X_{v} \sim \text{Multinomial}(p_{v,1}, p_{v,2}, p_{v,3}, p_{v,4}, p_{v,5})
\end{split}
\end{equation}

To summarize, all the above components form a modular representation of VSR morphogenesis. The general task requirements are distributed to each position as voxel-specific functionalities, which are then coordinated holistically through message passing and decoded into a cohesive morphological structure. Such a modeling approach affords us a reduced-rank problem formulation, while explicitly enhancing the learning of sophisticated inter-voxel dependencies. The conditional generative modeling additionally enables us to exploit the structural similarities underlying multiple tasks and facilitates a transfer of design experience via parameter sharing (please see Supplementary Material C for empirical evidence of inter-task knowledge sharing). This problem-oriented neural architecture design and multi-task optimization scheme together foster a more sample-efficient VSR design process, as verified by our experimental results. 

\subsection{Fitness evaluation}
\label{sec:Control optimization and fitness evaluation}

\subsubsection{Control optimization}

Once the robot morphologies are generated, the next step involves training a senserimotor control policy for each of them so as to evaluate their task completion abilities. In this work, we employ the Proximal Policy Optimization (PPO) algorithm \cite{schulman2017proximal} for control optimization. The control policy \(\pi(a_t | o_t)\) acts as a mapping from environmental observations $o_t$ to actuation signals $a_t$. Equipped with an optimized policy, the robot can choose appropriate actions and effectively interact with its environment. The cumulative reward obtained during these interactions then serves as a measure of the robot's fitness, indicating how qualified it is for a given task. 

During each iteration of control optimization, the robot interacts with the environment and collects trajectories comprising observations, actions, and rewards. Specifically, at each time step, the robot receives an observation $o_t$, selects an action $a_t$ according to its current policy, executes the action, and receives a reward $r_t$ from the environment. The trajectory keeps rolling out until a termination state or a pre-specified time step limit is reached. These collected trajectories are then used to update the parameters in the policy network so that it assigns higher probabilities to rewarding actions. 

PPO enhances conventional policy-gradient algorithms by introducing importance sampling, which allows for offline policy updates using old trajectories and thereby improves sample efficiency. As shown in equation (\ref{eq:ppo}), the PPO objective function consists of two terms. The first term encourages the policy to take actions with greater advantage. The second term constrains the step size of policy updates using Kullback-Leibler divergence, preventing excessively large changes that could destabilize the training process. For further details on the PPO algorithm, please refer to \cite{schulman2017proximal}. 

\begin{equation}
\label{eq:ppo}
\begin{split}
\underset{w}{\text{Maximize}} \quad
\Biggl\{ \ 
& E_{t} \left[ \frac{\pi_w(a_t|s_t)}{\pi_{w_{\text{old}}}(a_t|s_t)} \hat{A}_t \right] \\
& - \beta \times 
KL\bigl[\pi_w(\cdot|s_t), \pi_{w_{\text{old}}}(\cdot|s_t)\bigr]
\Biggr\},
\end{split}
\end{equation}
where $w$ denotes the trainable parameters in the policy network, $\pi_w$ and $\pi_{w_{\text{old}}}$ represent the policy being updated and the policy used for collecting trajectories, respectively. $\hat{A}_t$ is the advantage estimator of actions, such as the Generalized Advantage Estimator (GAE). \(\beta\) is a hyperparameter controlling the step size of updates.

\subsubsection{Construction of the global morphology-fitness pool}

After control policies are optimized, each robot is reintroduced into the simulation environment to interact until reaching a terminal state, thus forming a complete trajectory. The cumulative reward of this trajectory serves as the fitness evaluation $f$ of the robot morphology in the corresponding task environment. The morphology, along with its fitness, is stored in a sample pool referred to as the \emph{global morphology-fitness pool}. This sample pool accumulates evaluated robot designs in each generation, facilitating sample reuse across different evolutionary stages as well as different task environments and thus greatly boosting sample efficiency. 

\subsection{Natural selection}
\label{sec:A-B balance}
In each generation, a collection of samples is drawn from the morphology-fitness pool through a natural selection operator. The samples are then used for parameter updates of MEC-VAE. However, to improve upon traditional evolutionary algorithms that solely preserve the elite and better resemble evolution in nature, we introduce a degree of stochasticity into the sampling process. 

Concretely, we argue that traditional EAs are more or less misguided by the principle of "\emph{survival of the fittest}". Contrary to the "all or nothing" interpretation, natural selection is better understood as a probabilistic process, where certain traits are more likely, but not guaranteed, to be passed on to future generations \cite{gregory2009understanding}. As evidence, \cite{beardmore2011metabolic} demonstrates through experiments that, under specific environmental conditions, both fit and unfit bacteria can coexist, resulting in stable diversity. Consequently, the complete rejection of suboptimal solutions in conventional selection operators is likely to result in limited evolutionary directions, thereby hindering biodiversity. 

To this end, we employ a two-step procedure to select $n$ robot morphologies from each task. We begin with sorting the global morphology-fitness pool of each task in descending order of fitness, and select the top $n\times\alpha$ morphologies. This step focuses on preserving or exploiting advantageous morphologies. The remaining $n\times(1-\alpha)$ morphologies are then randomly sampled from the pool, with probabilities positively correlated with fitness (as shown in equation (\ref{eq:softmax_used})). This serves as an enhancement of diversity. The hyperparameter $\alpha$ controls the degree of preference for elites. A higher $\alpha$ would render the evolutionary process more exclusively favoring top-ranking robot designs, while a lower $\alpha$ would increase the randomness of elitism selection, allowing more suboptimal robot designs to be explored. By adjusting $\alpha$, we can strike a flexible balance between exploitation of \textbf{a}dvantage and exploration of \textbf{b}iodiversity, and thus we term this sampling technique as A-B balancing. 

\begin{equation}
\label{eq:softmax_used}
P_{\text{chosen}}(X_i^t) = \frac{\exp(f_i^t)}{\sum_{j=1}^{I}\exp(f_j^t)}, \quad i = 1, 2, \ldots, I \quad 
\end{equation}
where \( X_i^t \) denotes the \( i \)-th robot in the morphology-fitness pool of task \( t \), $f_i^t$ denotes its fitness, and \( I \) is the total number of evaluated robots for each task. 

\subsection{Model update}
\label{sec:model update}

Finally, with the selected $n$ morphologies for each task, we update MEC-VAE through maximum likelihood estimation (MLE). However, since MEC-VAE involves latent variables in its generative process, the marginal likelihood function of observable variables (\emph{i.e.,} the selected morphologies and their tasks) becomes intractable, making exact solutions impractical. To address this challenge, we resort to variational inference, an effective technique for training probabilistic generative models. By employing Jensen's inequality, we derive a lower bound of the original likelihood function (equation (\ref{eq:elbo})), known as the Evidence Lower Bound (ELBO): 

\begin{equation}
\label{eq:elbo}
\begin{split}
\textit{ELBO} = & \ E_{t \sim \mathcal{T}} E_{X \sim \mathcal{X}_t} \biggl\{ 
E_{H_t \sim q_{\phi}(H_t|X,t)}\bigl[\log p_{\theta}(X|H_t)\bigr] \\
& \quad - D_{KL}\bigl(q_{\phi}(H_t|X,t) \,||\, p_{\theta}(H_t|t)\bigr)
\biggr\},
\end{split}
\end{equation}

where $\mathcal{T}$ denotes a uniform distribution over all tasks, and $\mathcal{X}_t$ denotes the elite samples of task $t$ as explained in the previous subsection. $p_\theta$ and $q_\phi$ stand for the generative process and the approximate posterior, respectively.  

The derivation of ELBO, as detailed in Supplementary Material B, requires an approximate posterior distribution $q_{\phi}(H_t|X,t)$. It is also referred to as the encoder of a VAE, which infers the distribution of latents from observable variables. We construct this approximate posterior using a pair of MLPs. Specifically, these MLPs both take the robot morphology $X$ and the corresponding task type $t$ as input, and compute the mean vector and covariance matrix of the task latents $H_t$, respectively. We denote these MLPs as $\mu_{\phi}(X,t)$ and $\Sigma_{\phi}(X,t)$. Consequently, the approximate posterior distribution of $H_t$ is expressed as $\mathcal{N}(\mu_{\phi}(X,t),\Sigma_{\phi}(X,t))$

As seen in equation (\ref{eq:elbo}), the ELBO of MEC-VAE consists of two terms. The first is the reconstruction loss, which measures the similarity between the original samples and the reconstructed ones after encoding and decoding. The second is a regularization term that penalizes the divergence between the approximate posterior and the prior distribution of latent variables.

It can be shown that, assuming sufficiently high-capacity approximate posterior, maximizing the Evidence Lower Bound is equivalent to maximizing the likelihood function. We perform multiple steps of gradient ascent in each generation with respect to the ELBO to update MEC-VAE. However, the calculation of ELBO involves a sampling operation, $H_t\sim q_{\phi}(H_t|X,t)$. Direct sampling would prevent gradients from propagating through the encoder. To address this problem, we utilize the reparameterization trick. By sampling noise \(\epsilon\) from \(\mathcal{N}(0, I)\) and reformulating the sampling process as \(H_t = \epsilon \odot \Sigma_{\phi}(X,t) + \mu_{\phi}(X,t)\), we enable gradient propagation through $\mu_{\phi}(X,t)$ and $\Sigma_{\phi}(X,t)$. 

\begin{figure*}[htbp]
    \centering
    \includegraphics[width=1\textwidth]{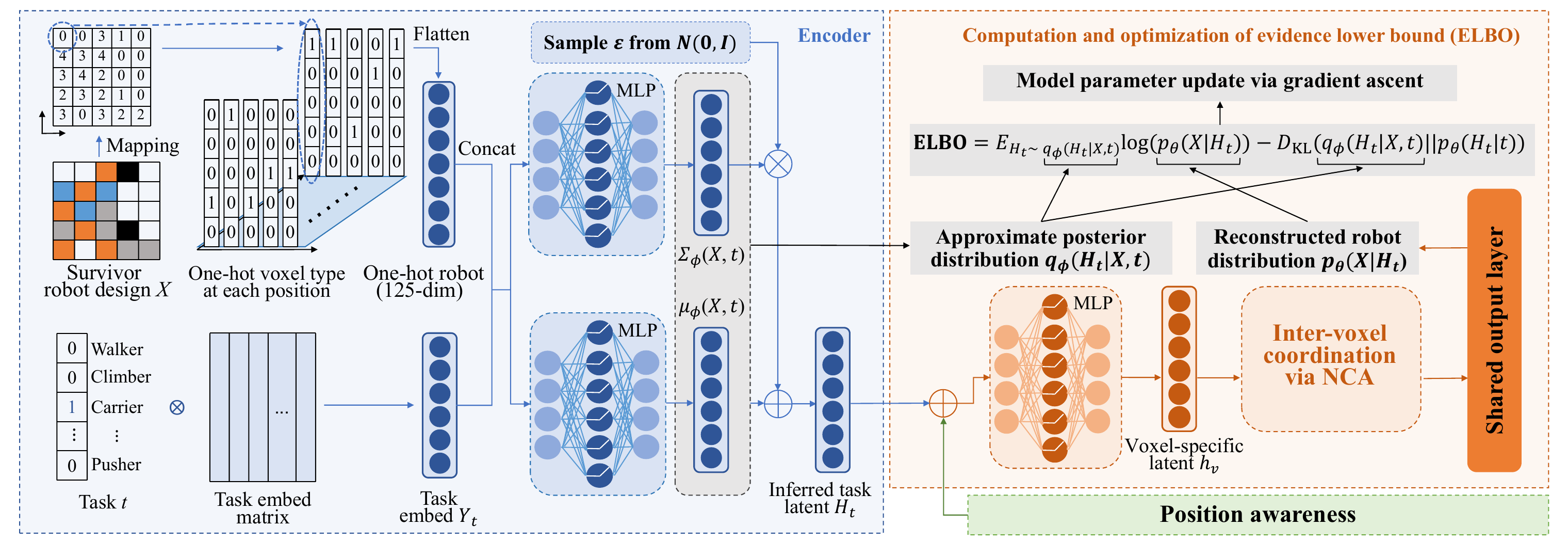}
    \caption{Illustration of model update via variational inference. }
    \label{fig:posterior}
\end{figure*}

The specific procedure of model update via variational inference is illustrated in Fig. \ref{fig:posterior}. Taking surviving robot morphologies $X$'s and their associated task types $t$'s as inputs, the encoder constructs approximate posterior distributions of task latents, $q_{\phi}(H_t|X,t)$, which take the place of the prior distribution $p_{\theta}(H_t|t)$ in the robot generative process to yield \emph{reconstructed} distributions of robot designs. We then construct an evidence lower bound of the likelihood of surviving morphologies (\emph{i.e.}, ELBO) that consists of two components: a reconstruction error that measures how well the reconstructed distributions match the surviving morphologies, and a Kullback-Leibler divergence term that regularizes the approximate posteriors towards the prior distribution. ELBO then serves as the (negative) loss function to guide the parameter update of MEC-VAE. 

\begin{algorithm}[t!]
\caption{Pseudocode of MISCO}
\label{Algorithmic pseudo-code}
\KwIn{Number of tasks $T$, number of generations $G$, number of robot samples per update $n$, the proportion of advantage samples $\alpha$, the number of NCA steps $K$, \textit{population size} and the number of VAE updates per generation (\textit{train step}).}
\KwOut{Optimized parameters in MEC-VAE, evolved robot designs for each task.}

\For{$g=1, 2, \dots, G$}{
    \textbf{Step1} Generate Robots by Sampling from MEC-VAE \\
    \For{$t=1, 2, \dots, T$}{
        $\mathbf{Y}_t \gets$ Task Embedding Layer $(t)$;
        $\mu_t \gets$ $\mu_\theta(\mathbf{Y}_t)$, $\Sigma_t \gets$ $\Sigma_\theta(\mathbf{Y}_t)$\\
        \For{$j=1, 2, \dots, \textit{population size}$}{
            $\mathbf{H}_t \gets$ Sample Latent Task Vector from $\mathcal{N}(\mu_t, \Sigma_t)$\\
            \For{each position $\mathbf{v}$ in robot $\mathbf{X}_{t,j}$, $\textit{v} \in \{1, 2, \dots, 5 \times 5\}$}{
                $\mathbf{P}_v \gets$ Position Embedding Layer(Coordinate$(\mathbf{v})$)\\
                $\mathbf{h}_v \gets$ Concat$(\mathbf{H}_t, \mathbf{P}_v)$\\
                $\mathbf{H}_v \gets$ \textbf{NCA}$( \mathbf{h}_v, \mathbf{h}_{\textit{left}}, \mathbf{h}_{\textit{right}}, \mathbf{h}_{\textit{up}}, \mathbf{h}_{\textit{down}})$,  \textit{for $K$ times}, \text{with} \textit{left}, \textit{right}, \textit{up}, \textit{down} $\in$ neighbor voxels of \textit{v}\
                $\mathbf{(p_{v,1}, p_{v,2}, \dots, p_{v,5}}) \gets$ $F_\theta(\mathbf{H}_v)$\; 
                $\mathbf{X}_v \gets$ Sample Voxel Type from Multinomial$(p_{v,1}, p_{v,2}, \dots, p_{v,5})$\
            }
        }
    }
    \textbf{Step2} Optimize Controllers for Robots to Evaluate Fitness \\
    \For{$t=1, 2, \dots, T$}{
        \For{$j=1, 2, \dots, \textit{population size}$}{
            Optimize controller $C_{t,j}$ of robot design $\mathbf{X}_{t,j}$ by PPO\\
            Roll out a complete episode with $(\mathbf{X}_{t,j}, C_{t,j})$ in simulation environment to evaluate fitness\\
            Save design, controller, along with their fitness, into the morphology-fitness pool\
        }
    }
    {\textbf{Step3 \& 4} Update MEC-VAE via Variational Inference and A-B Balancing} \\
    \For{$i=1, 2, \dots, \textit{train step}$}{
        For each task, choose the top $n \times \alpha$ robots, and then sample $n \times (1 - \alpha)$ robots with replacement based on probabilities defined in equation (\ref{eq:softmax_used}), from the morphology-fitness pool\\
        Do one step of Stochastic Gradient Ascent towards maximizing the ELBO defined in equation (\ref{eq:elbo}), based on the selected $n \times T$ elite robots
    }
}
\end{algorithm}

\subsection{Theoretical guarantees of convergence}
\label{sec:Theoretical guarantees of convergence}
In this section, we establish theoretical guarantees for the convergence of our proposed method. We first demonstrate that MISCO asymptotically converges to global optima, and then show that it exhibits linear convergence in terms of the average convergence rate. The proofs of these theorems are relegated to the Supplementary Material A. 

\subsubsection{Asymptotic convergence to optimal solutions}
Let $\theta_\tau$ denote the parameters of MISCO in generation $\tau$, and let $p_{\theta_\tau}(x;t)=\int p_{\theta_\tau}(H_t|t)\cdot p_{\theta_\tau}(x|H_t)\text{d}H_t$ denote the robot design distribution of task $t$ learned by VAE. Define the entire robot design space as $\mathcal{X}$, and any specific design within this space as $x$. Also denote the task set as $\mathcal{T}$. Note that given the highly non-convex and multi-modal nature of VSR fitness landscape, we do not assume that the optimal robot design for each task $t$ is unique. Instead, we allow for multiple globally optimal designs, collectively denoted as $\mathcal{A}(t)$. Denote the distribution of elite samples (i.e. robots selected from the morphology-fitness pool to fit the VAE model) for task $t$ in generation $\tau$ as $C_{\tau}(x;t)$. The updating process of MEC-VAE can thus be approximated by a weighted average between the previous distribution and the elite distribution: $p_{\tau}(x;t)=(1-\alpha_\tau)p_{\tau-1}(x;t)+\alpha_\tau C_{\tau}(x;t)$, for $x\in \mathcal{X}$ and $t\in\mathcal{T}$ (also known as \emph{smoothed updating}). We now show that MEC-VAE asymptotically converges to the optimal solutions under mild conditions. \\

\textbf{Theorem 1 (asymptotic convergence of MEC-VAE)} Consider the evolutionary process of MEC-VAE as a stochastic process, with its state represented as $(p_\tau(t), X_{\tau}(t))\quad(\forall t\in\mathcal{T})$. Here $p_\tau(t)$ denotes the distribution defined by the generative process of VAE for task $t$ in generation $\tau$, and $X_{\tau}(t)$ stands for the set of elite solutions of task $t$ in generation $\tau$ (i.e. the support of $C_{\tau}(x;t)$). Assume the aforementioned sequence $\alpha_t$ satisfies:
\begin{equation}
    \label{eq:alpha1}
    \alpha_\tau\leq 1-\frac{\log(\tau+1)}{\log(\tau+2)},\quad (\tau\geq T) 
\end{equation}
for some $T>0$, and
\begin{equation}
    \label{eq:alpha2}
    \sum_{\tau=1}^{\infty}\alpha_\tau=\infty.
\end{equation}
It then follows that\\
\textbf{(a)} $X_{\tau}(t)$ converges to $X^*(t)$ with probability one as $\tau\rightarrow\infty$, where $X^*(t)$ represents an arbitrary set of robot designs that include at least one optimal solution of task $t$. \\
\textbf{(b)} $p_\tau(t)$ converges to $p^*(t)$ as $\tau\rightarrow\infty$, where $p^*(x;t)$ is an \emph{optimal distribution} that assigns non-zero probability only to optimal designs. In other words, the distribution learned by MEC-VAE would eventually degenerate onto optimal solutions. Note that here we do not discriminate $p^*$'s with different probability allocation on $\mathcal{A}$, but rather treat them equally as optimal distributions. 

\textbf{Theorem 1} examines the asymptotic behaviour of MEC-VAE, showing that it will not only find at least one optimal solution almost surely, but also eventually converge to a probability distribution with non-zero mass only on optimal solutions. We now proceed to investigate the convergence speed, specifically the average convergence rate (ACR), of MEC-VAE. 

\subsubsection{Linear convergence rate}
Before delving into the analysis of ACR, we first provide its formal definition, originally proposed in \cite{chen2021average}. 

\textbf{Definition 1 (average convergence rate, ACR)} Consider the maximization problem: $$\max{f(\textbf{x})},\quad \textbf{x}=(x_1,\cdots,x_d)\in \mathcal{S}\subset  \mathbb{R}^d, $$
where $f(\cdot)$ is the objective function, $\mathcal{S}$ is the domain of feasible solutions, and $f^*$ denotes the maximal value of $f$. Let $X_\tau$ be the population at generation $\tau$, with its fitness defined as $f(X_\tau):=\max\{f(x)|x\in X_\tau\}$, and its discrepancy with $f^*$ as $e(X_\tau):=f^*-f(X_\tau)$. The {\emph{average convergence rate}} (ACR) for $\tau$ consecutive generations of an evolutionary algorithm, starting from the initial population, is then defined as: 
\begin{equation}
\nonumber
ACR_\tau:=1-\left( \frac{e_\tau}{e_0}\right)^{1/\tau}=1-\left(  \prod_{k=1}^\tau CR_k \right)^{1/\tau}, \quad \tau\in \mathbb{Z}^+, 
\end{equation}
where $e_\tau:=\mathbb{E}[e(X_\tau)]$ and $CR_k:=e_k/e_{k-1}$.\\

From \textbf{Definition 1}, we see that a larger $ACR$ indicates faster convergence. $ACR=1$ means the optimal solution has been found, while $ACR<0$ suggests a decline in performance compared to $X_0$. {\emph{Linear convergence}}, defined as $\lim_{\tau\rightarrow +\infty}ACR_\tau=C>0$, is considered a desirable convergence rate. \\

\textbf{Definition 2 ($\rho$-promising region)} The \emph{$\rho$-promising region} of a population $X$ is defined as $S(X,\rho):=\{ Y\subset \mathcal{S} | e(Y)<\rho e(X)  \}$, i.e., all the populations that improve upon $X$ by a factor of $\rho$. With $\rho=1$, $S(X,\rho)$ is simply referred to as a promising region. \\

\textbf{Theorem 2 (linear convergence of ACR)} For each specific task $t\in\mathcal{T}$, let $P^{(\kappa)}(X(t);Y(t))$ denote the probability of transitioning from population $X(t)$ to population $Y(t)$ over $\kappa$ generations during evolution. Here a \emph{population} is defined as the training samples selected from the morphology-fitness pool in each generation. Assume MEC-VAE satisfies $$C_\rho:=\inf \{ P^{(\kappa)}(X(t);S(X(t),\rho)); X(t)\neq X^*(t)\}>0$$ for some $0<\rho<1$ and positive integer $\kappa$. This assumption implies that the probability of transitioning from a non-optimal population to one of its $\rho$-promising region over $\kappa$ generations is bounded below by a positive value. Then, we have that $\lim_{\tau\rightarrow \infty}ACR_\tau \geq C>0$, i.e., MEC-VAE achieves linear convergence. \\

\textbf{Theorem 2}, together with \textbf{Theorem 1}, indicates that MEC-VAE is not only guaranteed to asymptotically converge to optimal solutions, but also enjoys a rapid convergence speed so that it could hopefully reach optimal solutions (or at least near-optimal solutions) within limited generations.

\section{Results}
\subsection{Experiment settings}
We examine MISCO on a comprehensive suite of tasks selected from EvoGym. These tasks, as shown in Fig. \ref{fig:best fitness}(a), span various locomotive and manipulative tasks with all levels of difficulty. Detailed descriptions of these tasks are provided in Supplementary Material D. We draw comparisons against both classical evolutionary algorithms, including GA \cite{michalewicz2013genetic}, BO \cite{kandasamy2018parallelised} and CPPN-NEAT \cite{stanley2007compositional}, and state-of-the-art deep-GM assisted EDAs, including RoboGAN \cite{hu2022modular} and MorphVAE \cite{song2024morphvae}. Please see Supplementary Material E for a detailed introduction to these baselines. 

To facilitate a fair comparison across different methods, we run each of them until 1,000 robot designs have been evaluated. The population size is set as 25. All experimental results are averaged across three independent runs to reduce randomness. For further implementation details please refer to Supplementary Material G. We also conduct comprehensive ablation studies to validate the essentiality of MISCO's key components, with results reported in Supplementary Material F. 

\subsection{Two variants of MISCO}

Based on different preferences for elite morphologies, we introduce two variations of MISCO: 

\begin{itemize}
    \item \textbf{MISCO-A that focuses on exploitation} \\ 
    This variant heavily reuses top-ranking morphologies for model updating, which account for 70\% of training sample. This configuration is intended to speed up the optimization process by concentrating on the most advantageous morphologies. It is thus named \textbf{MISCO-A}, where "A" stands for \emph{advantage}. 
    \item \textbf{MISCO-B that focuses on exploration} \\ 
    This variant is fitted exclusively on suboptimal morphologies randomly sampled from the population. This encourages broader exploration of the solution space in order to yield more diversified design options. This variant is thus named \textbf{MISCO-B}, where "B" stands for \emph{biodiversity}. 
\end{itemize}

\subsection{Analysis of evolutionary efficiency}
\label{sec:efficiency}
\subsubsection{Best fitness}

\begin{figure*}[!htb]
    \centering
    \includegraphics[width=0.9\textwidth]{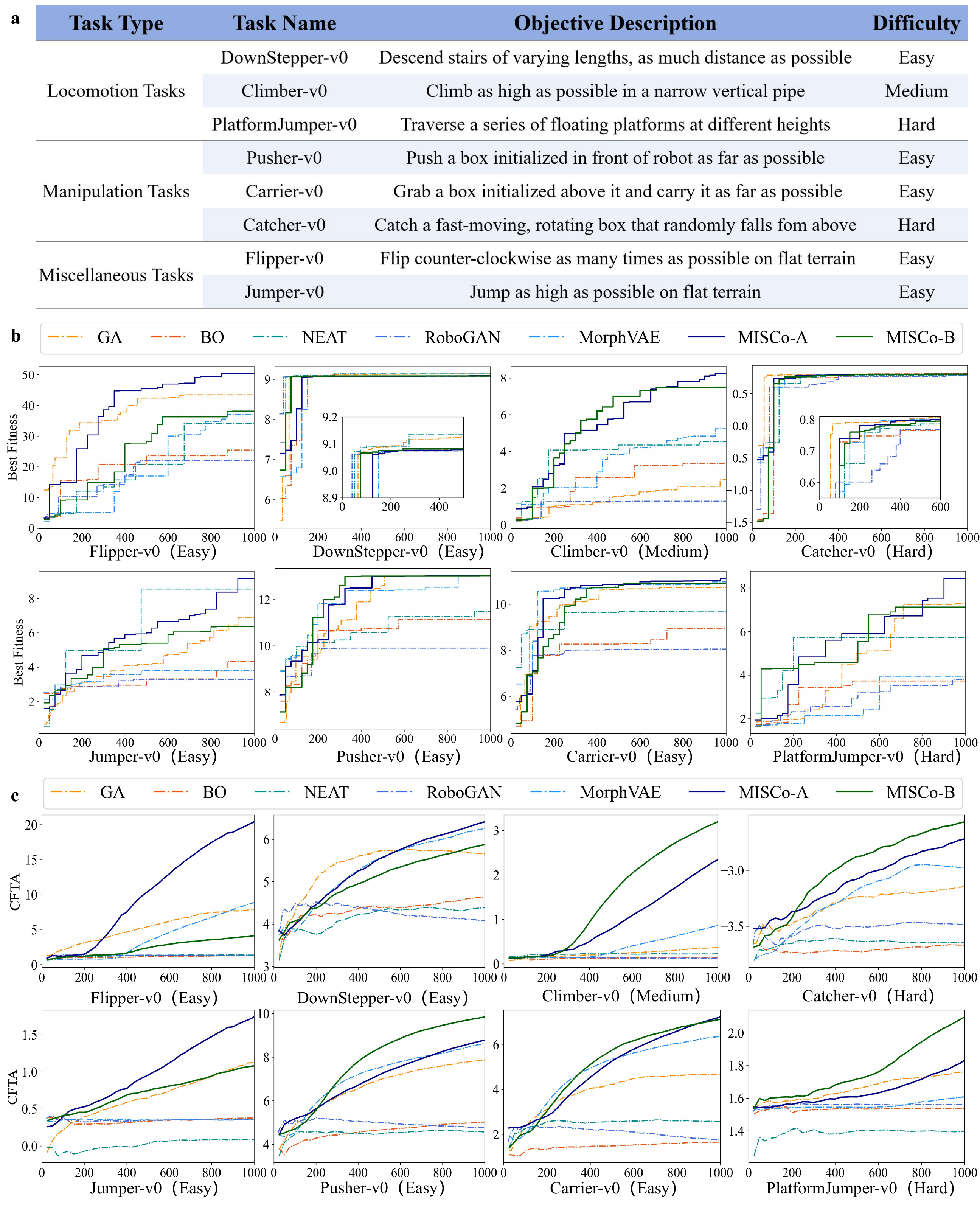}
    \caption{\textbf{Comparisons of best fitness and cumulative fitness time average (CFTA).} \textbf{a,} A brief introduction of the selected task instances. \textbf{b,} Comparison of best fitness. Best fitness (vertical axis) is plotted against the cumulative count of up to 1,000 robot evaluations (horizontal axis). \textbf{c,} Comparison of CFTA, which reflects how the average fitness of evaluated robot designs evolves over time. }
    \label{fig:best fitness}
\end{figure*}

\begin{figure*}[!hbt]
    \centering
    \includegraphics[width=\textwidth]{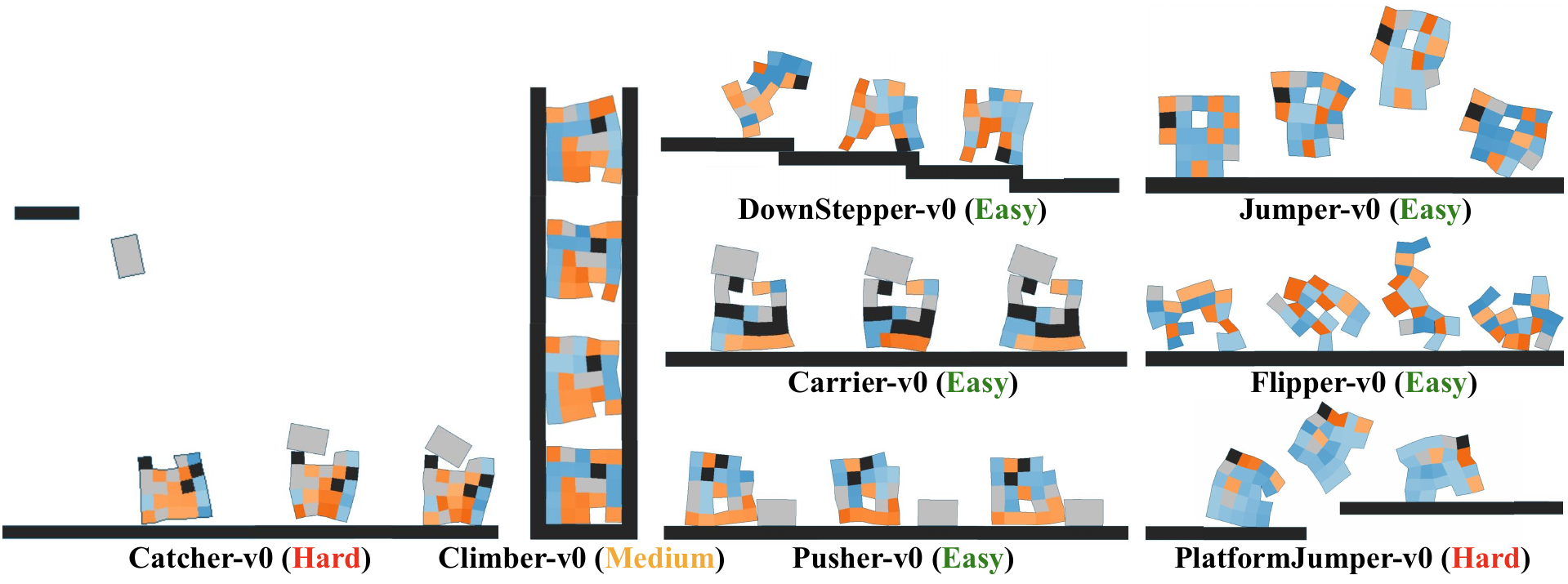}
    \caption{Best-performing morphologies evolved by MISCO. }
    \label{fig:visual}
\end{figure*}

We measure the fitness of the best-performing solution found within a specific number of evaluations. As shown in Fig. \ref{fig:best fitness} (b), MISCO-A achieves the best evolutionary outcomes in five out of eight tasks, while maintaining strong competitiveness in the remaining three tasks. Notably, no single baseline excels across all tasks. Although GA generally outperforms other baselines, its performance drastically deteriorates in specific tasks, such as Climber-v0, where it is among the least effective. In contrast, MISCO-A consistently yields superior robot designs with rapid convergence speed during morphological evolution. The best-performing robot morphologies evolved by MISCO are visualized in Fig. \ref{fig:visual}. 

We list the best fitness eventually achieved (\emph{i.e.,} at 1,000 evaluations) in Fig. \ref{fig:boxplot}(a) to more clearly showcase the superior optimization efficiency of MISCO-A. The highest results are marked in bold while the highest results among baselines are underlined. We further demonstrate the relative performances of baselines in a radar plot, where the best fitness achieved by each algorithm is divided by that of MISCO-A and reported as a percentage. We arrive at the following findings: (a) MISCO-A demonstrates highly robust evolutionary performances, ranking among the top three in all tasks and achieving the highest average rank. For the tasks where it does not come in first, such as DownStepper-v0, the best-performing baseline only exhibits \emph{marginal} advantage. (b) The potential of MISCO is unleashed in harder tasks that require more complex morphologies, such as Climber-v0 and PlatformJumper-v0. In these tasks, MISCO-A outperforms the baselines with \emph{significant} margins. (c) Additionally, MorphVAE also achieves commendable performance owing to its incorporation of variational autoencoders. But it underperforms MISCO in most cases, which highlights the importance of problem-oriented architectural design. 

The other variant, MISCO-B, demonstrates competitive performance across eight tasks. While it might slightly underperform certain baseline methods in a few cases, it maintains a well-balanced performance overall. Given the design of MISCO-B to emphasize exploration, it incorporates more randomness into natural selection, hence sacrificing some search efficiency compared to exploitation-focused methods like MISCO-A and GA. However, this makes it well-suited for problem scenarios that prioritize diversity. This distinct advantage of MISCO-B is discussed in subsection \ref{sec:diversity}.  

\subsubsection{Fitness distribution}

\textbf{Fitness distribution} refers to the distribution of fitness scores of all robot designs evaluated throughout evolution. It characterizes the overall performance of robot proposals over the evolutionary process, which complements the single-value summaries offered by best fitness and allows for more nuanced comparisons.   

\begin{figure*}[htbp]
    \centering
    \includegraphics[width=1\textwidth]{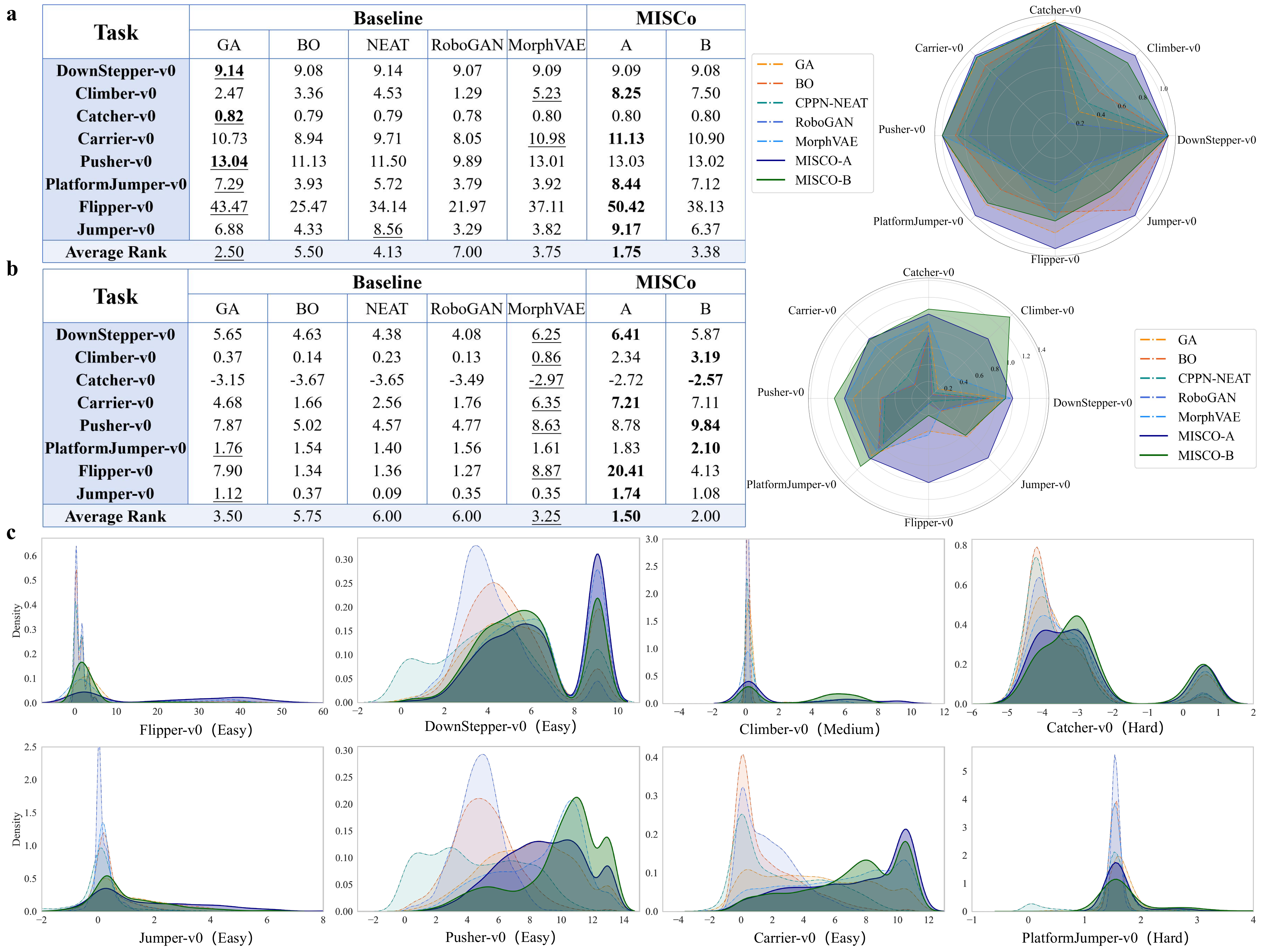}
    \caption{\textbf{Comparisons of evolutionary outcomes. a, left:} Comparative results of eventually achieved best fitness, as well as average ranks across all tasks. The best results are in bold while the best-performing baselines are underlined (the same is true for \textbf{panel b}). \textbf{a, right:} Normalized best-fitness performances showcased in radar plots, with MISCO-A serving as one unit. \textbf{b, left:} Comparative results of average fitness of all evaluated designs, as well as average ranks across all tasks. \textbf{b, right:} Normalized average-fitness performances showcased in radar plots, with MISCO-A serving as one unit. \textbf{c,} Kernel density estimations of fitness distributions. Probability density (vertical axis) is plotted against fitness (horizontal axis). }
    \label{fig:boxplot}
\end{figure*}

The kernel density estimations of different algorithms are plotted in Fig. \ref{fig:boxplot}(c). Notably, the fitness distributions of MISCO-A have more probability density located in high-fitness regions (reflected as the high peaks on the right), particularly in DownStepper-v0, Catcher-v0, Pusher-v0 and Carrier-v0. This indicates that MISCO-A not only excels in identifying optimal robot designs but also features a stronger directionality towards high-performance regions in the search space, leading to more stable evolutionary processes and fewer unsuccessful proposals. On the other hand, though shifting part of its emphasis to a broader range of exploration, MISCO-B still achieves fitness distributions that are superior or on par with baseline algorithms as well as MISCO-A. 

\subsubsection{Cumulative fitness time average (CFTA) }

\textbf{CFTA} refers to the cumulative average of fitness across generations. As shown in Fig. \ref{fig:best fitness}(c), MISCO-A demonstrates a steep and consistent increase in CFTA across all tasks, significantly outpacing baseline methods. Its focus on the exploitation of top-ranking designs allows it to not only gain an initial momentum towards high performance but also continuously refine these advantageous solutions to push the performance envelope towards optimality. This distinct behavior is largely enabled by the incorporation of MEC-VAE that explicitly tracks the distribution of high-performing robots, facilitating more targeted trial and error. The rapid and sustained growth in CFTA also reflects the design principle of MISCO-A, which focuses search efforts on exploitation and improvement of seen top solutions rather than exploration of potential ones. 

On the other hand, MISCO-B presents a different growing pattern in its CFTA curves. Unlike MISCO-A, this variant is instead designed to explore a broader range of solutions, which would initially cause a lag in CFTA. However, this exploration-driven approach pays off in later stages of evolution, when it uncovers a wider range of design options that would otherwise be overlooked and starts showing notable upward trends that surpass baselines. This distinct pattern of gradual discovery followed by a sharp increase underscores the strength of MISCO-B in identifying more evolutionary opportunities and building upon them to achieve efficient design optimization. The total average fitness (\emph{i.e.,} CFTA at 1,000 evaluations) of different algorithms and their relative performances are demonstrated in Fig. \ref{fig:boxplot}(b), where both MISCO-A and B show clear advantages. 

\subsection{Analysis of morphological diversity}
\label{sec:diversity}

Species diversity is how nature makes ecosystems robust to disruptive environmental changes. Likewise, a long-term goal of robotics is to develop diversified robot systems with high robustness in dynamic environments. Hence, we measure the diversity of evolved high-performing robot designs. Specifically, for each task we first pool together robot designs obtained by all different methods, and then calculate the 95\% quantile of their fitnesses. Robots with fitnesses that exceed this threshold are considered high-performing and involved in diversity calculation. 

We employ a metric named Morphological Diversity Score (MDS), which integrates Morphological Variance (MV) and Morphological Hamming Distance (MHD) into a unified measure. Concretely, MV assesses the overall variability in high-performing designs by calculating the entropy of material type for each voxel position. MHD, on the other hand, quantifies diversity at a finer granularity by calculating the Hamming distance between all pairs of high-performing robots. Thus, MDS is designed to capture both overall variability and pairwise distinctiveness, facilitating a comprehensive measurement of morphological diversity. For $5\times 5$ VSRs with 5 material types, their MDS is calculated as: 

\begin{equation}
\begin{split}
MDS =\ & 0.5 \times MV + 0.5 \times MHD \\
=\ & 0.5 \times \sum_{x=1}^{5} \sum_{y=1}^{5} \mathrm{Var}(D_{x,y}) \\
& + 0.5 \times \frac{2}{n(n-1)} \sum_{i=1}^{n-1} \sum_{j=i+1}^{n} \mathrm{Hamming}(M_i, M_j),
\end{split}
\end{equation}
where $D_{x,y}$ represents the voxel type distribution at position $(x, y)$ across all high-performing morphologies, and $\text{Var}(D_{x,y})$ is the variance of this distribution. \(n\) is the total number of high-performing morphologies. $M_i$ denotes a specific morphology. In this work, both factors are equally weighted to ensure a balanced consideration. 

\begin{table*}[htbp]
\centering
\caption{\textbf{Performance comparison of diversity.} In some cases the algorithm fails to produce more than one high-performing robot design, so diversity is not available ("NA"). }
\label{tab:algorithm performance of diversity}
\begin{tabular}{@{}c@{}}  
    \includegraphics[width=0.8\textwidth]{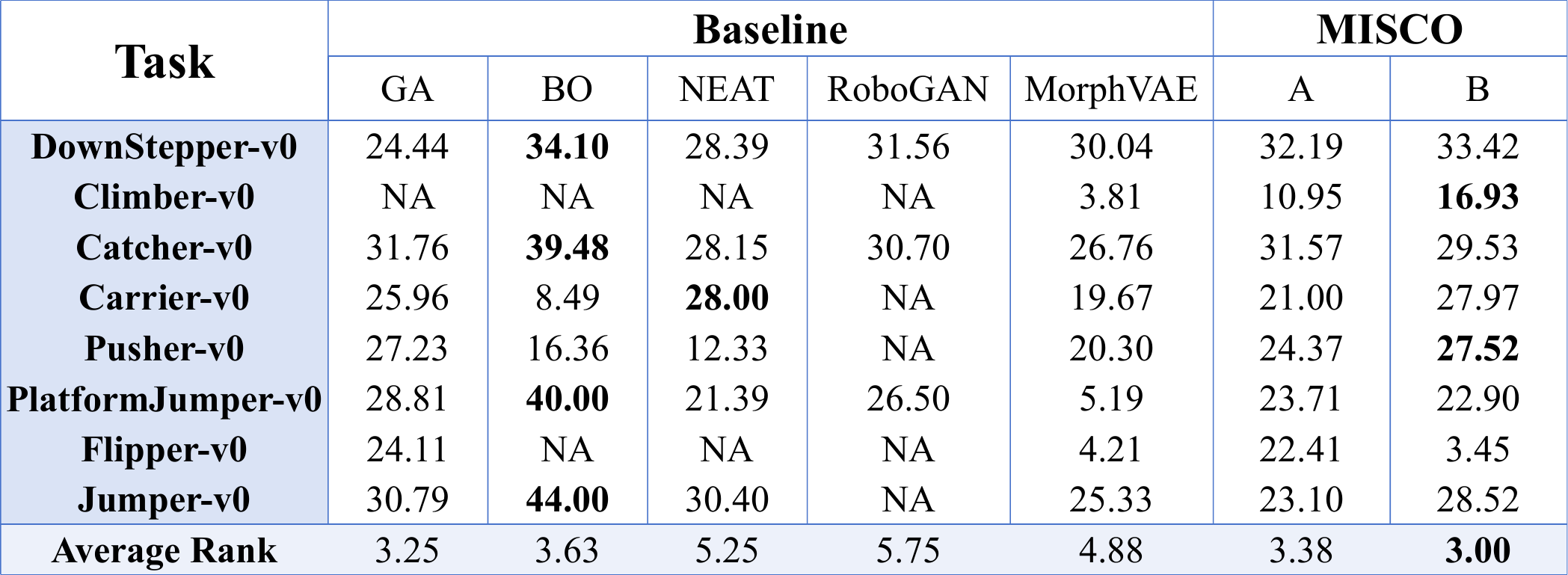}  
\end{tabular}
\end{table*}

As shown in Table \ref{tab:algorithm performance of diversity}, among all the methods evaluated, both MISCO variants exhibit notable robustness in producing diversified design options. The advantage of MISCO-B is particularly noteworthy with the highest average rank across different task settings. BO and GA also manifest strong competitiveness, but they compromise optimization efficiency, ending up among the least efficient algorithms in many cases. In contrast, MISCO strikes a favorable balance, which we attribute to the use of MEC-VAE to fit the distribution of elite samples, ensuring a sufficient coverage of solution spaces beyond scattered sample points. MISCO-B further encourages exploratory behaviors and achieve the highest diversity. These results highlight the under-explored potential of deep generative models to establish better exploration-exploitation trade-offs in optimization problems. Additionally, the performance gain of MISCO over MorphVAE demonstrates the benefit of carefully designed neural architecture, which better accounts for the intricate interaction dynamics within VSRs and possesses higher representational capacity. 

\section{Discussion}

We have presented MISCO, an innovative evolutionary framework for the design automation of voxel-based soft robots (VSRs). MISCO evolves VSRs by iteratively refining and sampling high-performing morphological distributions supported by a meticulously designed probabilistic generative model named MEC-VAE. Harnessing the representational capacity of variational autoencoders, MEC-VAE explicitly captures the intricate interaction dynamics among voxels across multiple task settings, hence supporting a precise characterization of the expressivity within VSR designs. We provide formal analysis regarding the convergence properties of MISCO, positioning it as a theoretically guaranteed and reliable algorithm for application. Through extensive simulated experiments, we demonstrate that MISCO achieves superior optimization efficiency over baseline methods, with its advantage particularly pronounced in complex tasks. We further reveal that MISCO is not only proficient in identifying optimal designs within constrained computational budget, but also fosters stable evolutionary trajectories, manifested by a stronger directionality towards high-performance regions and judicious trail and error with fewer unsuccessful robot proposals. 

We provide additional insights into the nuanced exploration-exploitation dynamics of MISCO. We show that incorporating suboptimal robot morphologies into the training data can effectively stimulate exploratory behaviors from MEC-VAE and achieve leading performance in morphological diversity without significantly sacrificing optimization efficiency. This reveals an appealing possibility of deep GM-assisted EDAs to support adaptive evolutionary strategies and accommodate various requirements in robotic system development. 

Regarding future directions, we expect our work to inspire studies that tackle extensive optimization challenges with similar deep GM-assisted evolution and dedicated neural architectures. We also find it intriguing to investigate how our meticulously designed generative model could enhance the interpretability of soft robot design. For example, by analyzing the strengths of neuron connections within inter-voxel message passing, we could gain additional insights into the operational mechanisms of VSRs. Furthermore, given the recent advances in "universal control" that highlight modular decomposition for sample-efficient control policies \cite{gupta2022metamorph}, \cite{hao2024heteromorpheus}, another promising direction would be to exploit the inherent homogeneity between design and control in modular robotic systems, allowing deep generative models with modular representations to serve dual purposes and achieve closed-loop development of embodied intelligence. 

\section*{Funding Declaration}
This research is supported by Intelligent Game and Decision Laboratory, National Natural Science Foundation of China (72371241), the MOE Project of Key Research Institute of Humanities and Social Sciences (22JJD910001), the Big Data and Responsible Artificial Intelligence for National Governance, Renmin University of China, Public Computing Cloud, Renmin University of China, and Zhiqiang Foundation.


\appendices

\section*{\textbf{Supplementary Material A:} \\Proofs of \textbf{Theorems 1} and \textbf{2}}
\label{appendix:proofs}
\textbf{\emph{proof of Theorem 1: }}The theorem is proved for $\forall t\in\mathcal{T}$, and we omit $t$ from now on for brevity. Consider the probability that the optimal solutions are never obtained:
     
\begin{equation}
\begin{split}    
\label{eq:never}    
&P\{X^* \text{ is never obtained }\} \\    
&=\prod_{\tau=0}^\infty P\{X^*\text{ not obtained in generation }\tau| \\    
&\quad X^*\text{ not obtained before generation }\tau\}.     
\end{split}
\end{equation}

We can calculate a lower bound of $p_{\theta_\tau}(x)$, which corresponds to the case when $x$ didn't show up as an elite sample through all previous generations:
\begin{equation}
\nonumber
\begin{split}
\label{eq:bound}
p_{\theta_\tau}(x)&\geq \left[ \prod_{i=0}^{\tau}(1-\alpha_i)\right]p_{\theta_0}(x)\\ &\geq \left[ \prod_{t=0}^{T-1}(1-\alpha_i)\right]\left[\prod_{i=T}^{\tau}\frac{\log(i+1)}{\log(i+2)}\right]p_{\theta_0}(x)\\
&=\left[\prod_{i=0}^{T-1}(1-\alpha_i)\right]p_{\theta_0}(x)\frac{\log(T+1)}{\log(\tau+2)}=\frac{\text{const}}{\log(\tau+2)},
\end{split}
\end{equation}
where the second inequality follows from equation (10) in main text. Hence, the probability of VAE generating any of $\mathcal{A}$ in generation $\tau$ is greater or equal to $|\mathcal{A}|\cdot \text{const}/\log(\tau+2)$. Note that this lower bound holds for arbitrary scenarios that could happen before generation $\tau$, and therefore we have
\begin{equation}
\nonumber
\begin{split}
    &P\{X^*\text{ is not obtained in generation }\tau|\\ &X^*\text{ wasn't obtained before generation }\tau\}\\
    &\leq \left[1-\frac{|\mathcal{A}|\cdot \text{const}}{\log(\tau+2)}\right]^N,
    \end{split}
\end{equation}
where $N$ denotes the population size, and the inequality holds because individuals in the population are generated independently. Hence, the probability in Equation (\ref{eq:never}) is upper bounded by
$$\prod_{\tau=0}^{T-1}1\cdot\prod_{\tau=T}^\infty\left[ 1-\frac{|\mathcal{A}|\cdot \text{const}}{\log(\tau+2)} \right]^N=\prod_{\tau=T}^\infty\left[ 1-\frac{|\mathcal{A}|\cdot \text{const}}{\log(\tau+2)} \right]^N. $$
Further taking logarithm of the right-hand side, we have that
$$N\sum_{\tau=T}^\infty \log \left[ 1- \frac{|\mathcal{A}|\cdot \text{const}}{\log(\tau+2)}\right]\leq -N \sum_{\tau=T}^\infty \left( \frac{|\mathcal{A}|\cdot\text{const}}{\log(\tau+2)} \right)=-\infty,$$
where the inequality follows from the fact that $\log(x)<x-1$ for all $x\in(0,1)$, and the identity holds because $\sum_{\tau}(\log(\tau+1))^{-1}=\infty$. Consequently, $P\{X^* \text{ is never obtained}\}=0.$ Note that once one or more optimal solutions are obtained, they will remain in elite samples thereafter, and therefore $X_{\tau}$ converges to $X^*$ almost surely as $\tau\rightarrow\infty$. 

So far, we proved that at least one of the optimal solutions will be obtained with probability 1 as the number of generations tends to infinity. Let us denote by $\tau^*$ the generation in which $X^*$ is attained for the first time. Now, consider the probability of any $x$ that does not belong to $\mathcal{A}$ in generation $\tau^*+\Delta \tau$. To this end, notice that for any non-optimal $x$, due to the selection pressure favoring high-performing robot designs, once one or more optimal solutions are attained, there should exist $d<1$ such that $C_{\tau+1}(x)\leq d\cdot p_\tau(x)$. Hence, for any $x\notin\mathcal{A}$ and $\tau\geq \tau^*$, we have

\begin{equation}
\nonumber
\begin{split}
p_{\tau+1}(x)&=(1-\alpha_{\tau+1})p_\tau(x)+\alpha_{\tau+1} C_{\tau+1}(x)\\ &\leq(1-\alpha_{\tau+1}+d\alpha_{\tau+1})p_\tau(x),\quad \forall x\notin\mathcal{A}.
\end{split}
\end{equation}

Hence, for any $x$ that does not belong to the optimal solutions, we have
$$p_{\tau^*+\Delta \tau}(x)\leq \left[  \prod_{\tau=\tau^*+1}^{\tau^*+\Delta \tau}(1-\alpha_\tau+d\cdot \alpha_\tau) \right]\cdot p_{\tau^*}(x).$$
Since $\log \prod_{\tau=0}^\infty (1-(1-d)\cdot\alpha_\tau)=\sum_{\tau=0}^\infty \log (1-(1-d)\cdot\alpha_\tau)< -(1-d)\cdot\sum_{\tau=0}^\infty \alpha_\tau =-\infty$, where  "$<$" follows again from the fact that $\log(x)<x-1$ for $\forall x\in(0,1)$, and the second identity is due to equation (11) of main text, we know that $\prod_{\tau=0}^\infty (1-(1-d)\cdot \alpha_\tau)=0$. As a consequence, 

\begin{equation}
\begin{split}
\lim_{\Delta \tau\rightarrow \infty} p_{\tau^*+\Delta \tau}(x) & 
\leq \left[  \prod_{\tau=\tau^*+1}^\infty (1-(1-d)\cdot \alpha_\tau) \right] \cdot p_{\tau^*}(x) = 0,  \\
& \quad \quad \quad \quad \quad \text{for } \forall x \notin \mathcal{A},
\end{split}
\end{equation}
which means that the probability of any non-optimal design would tend to zero, and thus $p_\tau(x)$ converges to $p^*$. With this, we complete the proof of \textbf{Theorem 1}. \hfill $\square$\\

\textbf{\emph{Proof of Theorem 2: }}First, it is easy to see that $S(X_k,\rho)\subset S(X_k,1)$ for any $0<\rho<1$, and for any $Y\in S(X_k,\rho)$, we have $f(Y)-f(X_k)> (1-\rho)(f^*-f(X_k))$. Therefore, for $\kappa$ consecutive generations, we have: 
\begin{equation}
\nonumber
\begin{split}
&e(X_k)-e(X_{k+\kappa})\\
&=\sum_{Y\in S(X_k,1)}(f(Y)-f(X_k))P^{(\kappa)}(X_k;Y)\\
&> \sum_{Y\in S(X_k,\rho)}(1-\rho)(f^*-f(X_k))P^{(\kappa)}(X_k;Y)\\
&\geq (1-\rho)C_\rho e(X_k), 
\end{split}
\end{equation}
\noindent where the identity is due to the fact that we treat the elite samples as our population, and hence for $Y \notin S(X_k,1)$, either $P^{(\kappa)}(X_k;Y)=0$ or $f(Y)-f(X_k)=0$. 

Then, 
$$\frac{\Delta^{(\kappa)}e_k}{e_k}=\frac{\mathbb{E}[e(X_k)-e(X_{k+\kappa})]}{\mathbb{E}[e(X_k)]}> (1-\rho)C_\rho.$$
Consequently, it holds that
\begin{equation}
\nonumber
\begin{split}
\lim_{\tau\rightarrow+\infty}ACR_\tau&=\lim_{\tau\rightarrow+\infty}\left[  1-\left(  \prod_{k=1}^\tau \frac{e_k}{e_{k-1}} \right)^{1/\tau}\right] \\
&> 1-\left[  1-(1-\rho)\cdot C_\rho \right]=(1-\rho)C_\rho>0,
\end{split}
\end{equation}
which completes the proof of 
\textbf{Theorem 2}. \hfill $\square$\\

\section*{\textbf{Supplementary Material B:} \\Derivation of the evidence lower bound (ELBO)}
\label{appendix:elbo}
Here we present the derivation of our objective function ELBO.
$$
\begin{aligned}
log(p_{\theta}(X|t)) &= \int log(p_{\theta}(X|t))q_{\phi}(H_t|X,t) dH_t \\
&= E_{H_t \sim q_{\phi}(H_t|X,t)} log(p_{\theta}(X|t)) \\
&= E_{H_t \sim q_{\phi}(H_t|X,t)}log{\frac{p_{\theta}(X,H_t|t)}{p_{\theta}(H_t|X,t)}} \\
&= E_{H_t \sim q_{\phi}(H_t|X,t)}log{\frac{p_{\theta}(X,H_t|t)}{q_{\phi}(H_t|X,t)}} \\
& \quad \quad \quad \quad + E_{H_t \sim q_{\phi}(H_t|X,t)}log{\frac{q_{\phi}(H_t|X,t)}{p_{\theta}(H_t|X,t)}} \\
&= \textit{ELBO} + D_{KL}(q_{\phi}(H_t|X,t) || p_{\theta}(H_t|X,t)) \\
&\geq \textit{ELBO}
\end{aligned}
$$
Hence, 
$$
\begin{aligned}
\textit{ELBO} &= E_{H_t \sim q_{\phi}(H_t|X,t)}log{\frac{p_{\theta}(X,H_t|t)}{q_{\phi}(H_t|X,t)}} \\
&= E_{H_t \sim q_{\phi}(H_t|X,t)}[\log p_{\theta}(X|H_t)]  \\
& \quad \quad- D_{KL}(q_{\phi}(H_t|X,t) || p_{\theta}(H_t|t))
\end{aligned}
$$

\section*{\textbf{Supplementary Material C:} \\Empirical evidence of structures shared across tasks }
\label{appendix:shared_structures}
Here we present supplementary details of knowledge sharing for multiple tasks. As a preliminary study to elicit and support the design motivation of multi-task learning in our algorithm, we analyze the morphological structures of elite robot designs across multiple tasks, qualitatively and quantitatively, using dimensionality reduction visualization and perplexity calculations techniques. These elite morphological samples are sourced from the top 5\% of 1,000 robots evaluations from three separate single-task learning experiments—Genetic Algorithm (GA), Bayesian Optimization (BO), and CPPN-NEAT. Each experiment was conducted across multiple tasks and repeated three times to mitigate randomness, thus providing 450 elite samples per task and allowing a robust evaluation of shared structures across tasks. 

\begin{figure*}[!htb]
    \centering
    \includegraphics[width=1\textwidth]{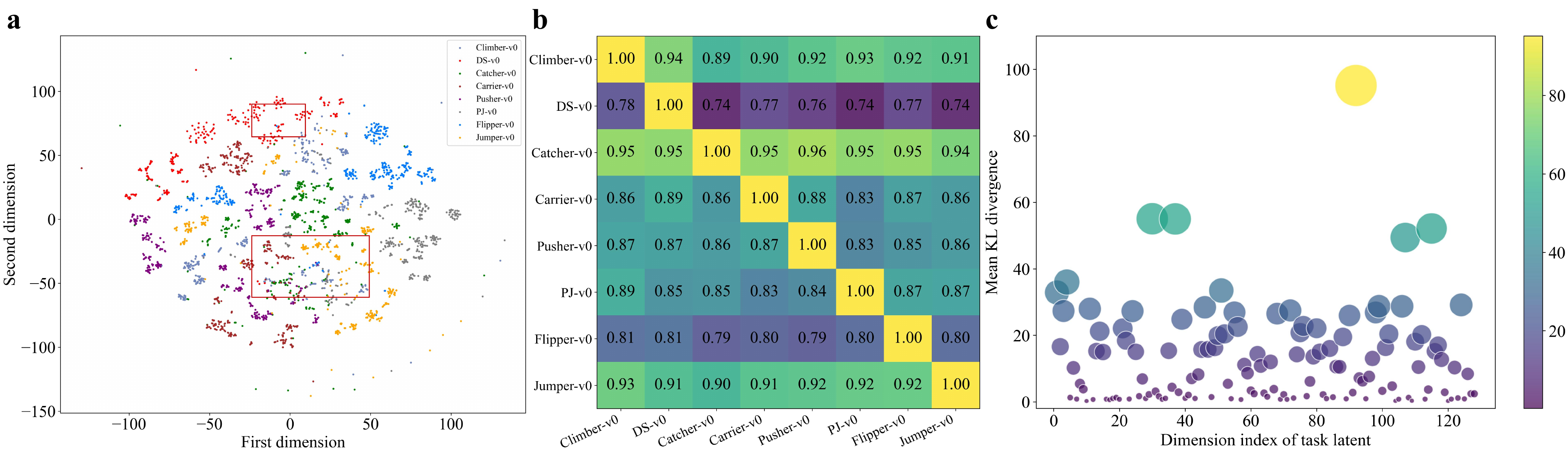}
    \caption{\textbf{Shared structures across tasks.} Note that 'DS-v0' and 'PJ-v0' are short for DownStepper-v0 and PlatformJumper-v0, respectively. \textbf{a,} Overlapping of robot designs in multiple tasks. \textbf{b,} Perplexity of robot designs in multiple tasks. \textbf{c,} Dimension-wise average KL divergence of task latent variables. The bubble plot illustrates the average KL divergence for each of the 128 dimensions of the task latent variable in our multi-task VAE framework. The $x$-axis shows the dimension index, ranging from 1 to 128, while the $y$-axis represents average KL divergence, with the bubble size and color intensity corresponding to divergence magnitude.}
    \label{fig: multi-task-intuitive}
\end{figure*}

As shown in Supplementary Figure \ref{fig: multi-task-intuitive} (a), the qualitative study leverages the $t$-SNE algorithm to map elite robots across various tasks into a two-dimensional space, effectively highlighting shared structures and potential synergies among different tasks. To be more specific, the overlap observed within the red boxes suggests that certain morphological features may serve multiple purposes, enabling cross-task functionality. These shared substructures could be leveraged to improve design efficiency, as they suggest underlying morphological similarities that can be beneficial across tasks. Consequently, multi-task optimization can realize knowledge sharing to improve sample utilization in morphological evolution of robots. 

To quantify the similarity or distinguishability between different task categories in the robot morphology, we compute the \textbf{perplexity}, a metric using the concept of intra-class and inter-class distances to measure how closely elite samples from different tasks are clustered or separated. The perplexity \( P_{ij} = \frac{d_{\text{within}}^i}{d_{\text{between}}^{ij}} \) represents the ratio of intra-class to inter-class distance, where values close to 1 indicate high similarity between tasks, and values less than 1 suggest significant separation. The intra-class average distance for a given task \( i \) measures the compactness within that task and is computed as the average of all pairwise distances among samples within \( i \): \( d_{\text{within}}^i = \frac{2}{N_i (N_i - 1)} \sum_{1 \leq m < n \leq N_i} \|\mathbf{x}_{im} - \mathbf{x}_{in}\| \), where \( N_i \) is the number of samples in task \( i \). The inter-class average distance between tasks \( i \) and \( j \) quantifies their separation and is calculated as \( d_{\text{between}}^{ij} = \frac{1}{N_i N_j} \sum_{m=1}^{N_i} \sum_{n=1}^{N_j} \|\mathbf{x}_{im} - \mathbf{x}_{jn}\| \), with \( N_j \) being the number of samples in task \( j \). 

Quantitative analysis demonstrated in Supplementary Figure \ref{fig: multi-task-intuitive} (b) reveals that all perplexity values are less than 1, but most are very close to 1. This pattern suggests a certain degree of independence among task categories: intra-class distances are consistently smaller than inter-class distances, particularly for tasks like "Walker-v0" and "DownStepper-v0". Such observations underscore their relative separation and the distinct substructures inherent to task-specific designs, making task awareness a crucial requirement. However, the proximity of most perplexity values to 1 with other tasks indicates a separable yet challenging-to-distinguish reality. This further validates the presence of overlapping features, likely arising from shared substructures that meet similar functional requirements across tasks, highlighting the importance of multi-task knowledge sharing. 

In conclusion, this preliminary analysis underscores the potential of multi-task optimization to promote knowledge-sharing across tasks. By reusing shared substructures, multi-task optimization could reduce the need for task-specific samples, thereby improving the overall efficiency and effectiveness of the evolutionary process. This knowledge-sharing approach fosters design adaptability across various robot tasks, expanding the utility and performance of evolved morphologies.

Next, we present an analysis of the information shared between multiple tasks in MEC-VAE. The bubble plot in Supplementary Figure \ref{fig: multi-task-intuitive} (c) visually represents the average KL divergence of each dimension of the task latent variable in a 128-dimensional Gaussian distribution. Each dimension corresponds to an independent component of the task latent space. The KL divergence of each dimension is averaged over all task pairs. Smaller bubbles and lower average KL divergence values indicate smaller divergence across tasks, suggesting that these dimensions contain shared information contributing to common functional substructures across tasks. Conversely, larger bubbles and higher KL divergence values reflect significant task-specific variation, implying that these dimensions encode unique characteristics or substructures for each task. This pattern highlights the model's ability for multi-task knowledge sharing: dimensions with low KL divergence facilitate shared representations, while dimensions with high KL divergence promote decoupling, thus achieving a balance between shared and task-specific information. These findings support the effectiveness of multi-task learning within MISCO in preserving both shared and distinctive task characteristics, aligning with the model's objective of simultaneously optimizing diversity and specialization (efficiency) at the task level. 

\section*{\textbf{Supplementary Material D:} \\Introduction to task settings}
\label{appendix:task-setting}
\begin{figure*}[htbp]
    \centering
    \includegraphics[width=1\textwidth]{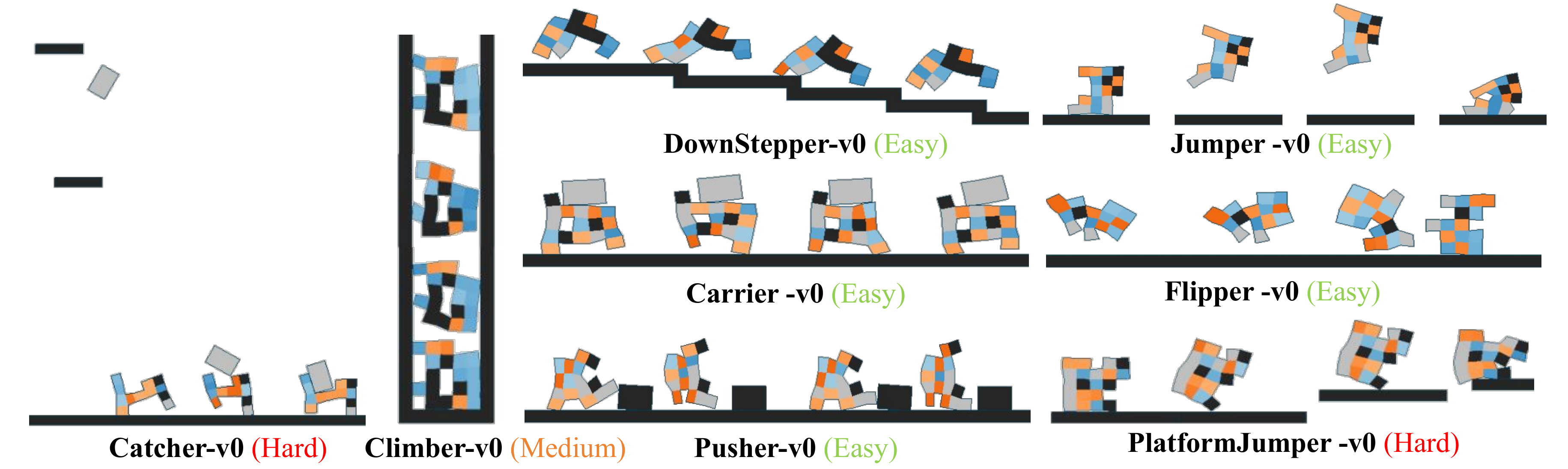}
    \caption{Illustration of various tasks in the EvoGym environment. }
    \label{fig:task_settings}
\end{figure*}

Here we provide a detailed introduction to the task settings, accompanied by illustrative images of each task (Supplementary Figure~\ref{fig:task_settings}). For readers seeking a deeper understanding of the physical setup and configurations, we recommend referring to \cite{bhatia2021evolution}.

\subsection{Climber-v0}

In this task the robot climbs as high as possible through a flat, vertical channel. This task is medium.  
Let the robot object be $r$. The observation space has dimension $S \in R^{n+2}$, where $n$ is the number of point masses in object $r$, and is formed by concatenating vectors $v^r, c^r$ with lengths 2 and $n$, respectively. The reward $R$ is  
\[
R = \Delta p^r_y,
\]  

which rewards the robot for moving in the positive $y$-direction.  


\subsection{DownStepper-v0}

In this task the robot climbs down stairs of varying lengths. This task is \textbf{easy}.  
Let the robot object be $r$. The observation space has dimension $S \in R^{n+14}$, where $n$ is the number of point masses in object $r$, and is formed by concatenating vectors $v^r, \theta^r, c^r, h_b^r(5)$ with lengths 2, 1, $n$, and 11 respectively. The reward $R$ is  

\[
R = \Delta p^r_x,
\]  

which rewards the robot for moving in the positive $x$-direction. The robot also receives a one-time reward of 2 for reaching the end of the terrain, and a one-time penalty of $-3$ for rotating more than 90 degrees from its original orientation in either direction (after which the environment resets).  

\subsection{Carrier-v0}

In this task the robot catches a box initialized above it and carries it as far as possible. This task is \textbf{easy}. Let the robot object be r and the box object the robot is trying to carry be $b$. The observation space has dimension $S \in R^{n+6}$, where n is the number of point masses in object r, and is formed by concatenating vectors $ v^b, p^b - p^r, v^r, c^r$ with lengths 2, n, 2, and 2, respectively. The reward $ R = R_1 + R_2 $ is the sum of several components. 
\[
R_1 = 0.5 * \Delta p^r_x + 0.5 * \Delta p^b_x,
\]

which rewards the robot and box for moving in the positive x-direction.

\[
R_2 = \begin{cases} 0 & \text{if} \ p^b_y \geq t_y, \\ 
10 * \Delta p^b_y & \text{otherwise}
\end{cases}
\]

which penalizes the robot for dropping the box below a threshold height $t_y$ .

\subsection{Pusher-v0}

In this task the robot pushes a box initialized in front of it. This task is \textbf{easy}. Let the robot object be $r$ and the box object the robot is trying to push be $b$. The observation space has dimension $S \in R^{n+6}$, where $n$ is the number of point masses in object $r$, and is formed by concatenating vectors $v^b, p^b - p^r, v^r, c^r$ with lengths 2, $n$, 2, and 2, respectively. The reward $R = R_1 + R_2$ is the sum of several components.  

\[
R_1 = 0.5 * \Delta p^r_x + 0.75 * \Delta p^b_x,
\]

which rewards the robot and box for moving in the positive $x$-direction. 

\[
R_2 = -\Delta |p^b_x - p^r_x|,
\]

which penalizes the robot and box for separating in the $x$-direction.  

\subsection{Catcher-v0}
In this task the robot catches a fast-moving, rotating box. This task is \textbf{hard}.  
Let the robot object be $r$ and the box object the robot is trying to throw be $b$. The observation space has dimension $S \in R^{n+7}$, where $n$ is the number of point masses in object $r$, and is formed by concatenating vectors $p^b - p^r, v^r, v^b, \theta^b, c^r$ with lengths 2, 2, 2, 1, and $n$, respectively. The reward $R = R_1 + R_2$ is the sum of several components.  

\[
R_1 = -\Delta p^b_x - \Delta p^r_x,
\]  

which rewards the robot for moving to the box in the $x$-direction.

\[
R_2 = 
\begin{cases} 
0 & \text{if} \ p^b_y \geq t_y, \\
10 \cdot \Delta p^b_y & \text{otherwise}
\end{cases}
\]  

which penalizes the robot for dropping the box below a threshold height $t_y$.  

\subsection{PlatformJumper-v0}

In this task the robot traverses a series of floating platforms at different heights. This task is \textbf{hard}. Let the robot object be $r$. The observation space has dimension $S \in R^{n+14}$, where $n$ is the number of point masses in object $r$, and is formed by concatenating vectors  

\[
v^r, \theta^r, c^r, h^r_b(5)
\]  

with lengths 2, 1, $n$, and 11 respectively. The reward $R$ is  

\[
R = \Delta p^r_x,
\]  

which rewards the robot for moving in the positive $x$-direction. The robot also receives a one-time penalty of $-3$ for rotating more than 90 degrees from its original orientation in either direction or for falling off the platforms (after which the environment resets).  


\subsection{Flipper-v0}
In this task the robot flips counter-clockwise as many times as possible on flat terrain. This task is \textbf{easy}. Let the robot object be $r$. The observation space has dimension $S \in R^{n+1}$, where $n$ is the number of point masses in object $r$, and is formed by concatenating vectors  

\[
\theta^r, c^r
\]  

with lengths 1 and $n$ respectively. The reward $R$ is  

\[
R = \Delta \theta^r
\]  

which rewards the robot for rotating counter-clockwise.  

\subsection{Jumper-v0}
In this task the robot jumps as high as possible in place on flat terrain. This task is \textbf{easy}. Let the robot object be $r$. The observation space has dimension $S \in R^{n+7}$, where $n$ is the number of point masses in object $r$, and is formed by concatenating vectors  

\[
v^r, c^r, h^r_b(2)
\]  

with lengths 2, $n$, and 5 respectively. The reward $R$ is  

\[
R = 10 \cdot \Delta p^r_y - 5 \cdot |\Delta p^r_x|,
\]  

which rewards the robot for moving in the positive $y$-direction and penalizes the robot for any motion in the $x$-direction.  

\section*{\textbf{Supplementary Material E:} \\Baselines}
\label{appendix:baselines}
We benchmark our proposed method, MISCO, against the following baselines: 
\begin{itemize}    
\item \textbf{Genetic Algorithm (GA)} \\
    The Genetic Algorithm \citep{michalewicz2013genetic} is an evolutionary algorithm inspired by natural selection. Solutions are first encoded as chromosomes in an GA. In each generation, the fittest solutions are selected for reproduction (through stochastic operations like mutation and crossover) in order to produce offspring for the next generation. 
    \item \textbf{Bayesian Optimization (BO)} \\
    The Bayesian Optimization algorithm \citep{kandasamy2018parallelised} is commonly used for black-box optimization, especially when the objective function is unknown or costly to compute. The Gaussian Process (GP) is usually employed as the surrogate model of the objective function. The Expected Improvement (EI) is one of the most common choices of the acquisition function that selects the next sampling points with a balance between exploitation and exploration. 
    \item \textbf{CPPN-NEAT} \\    
    CPPN-NEAT \citep{stanley2007compositional} is a predominant approach to the design automation of soft robots in past literature. It utilizes a Compositional Pattern Producing Network (CPPN) to parameterize each robot design, mapping voxel coordinates to material types. The NeuroEvolution of Augmenting Topologies (NEAT) algorithm is used to evolve the weights and architectures of these CPPNs, through a GA-like evolutionary process. 
    \item \textbf{RoboGAN} \\    
    RoboGAN \citep{hu2022modular} is one of the first approaches that leverage deep generative models for the design automation of modular robots. A generative adversarial network (GAN) is used to generate a set of robots at the beginning of each generation. These robots, treated as negative samples, then undergo several steps of GA evolution to yield positive samples. GAN is iteratively trained to produce robots that resemble positive samples, thus guided towards higher performance. 
    \item \textbf{MorphVAE} \\
    MorphVAE \citep{song2024morphvae} is arguably the first approach that employs a variational autoencoder (VAE) for the design automation of VSRs. While it has been proven to yield unprecedented optimization efficiency, its VAE consists of simple multi-layer perceptrons without carefully designed neural architectures, leaving much of the generative models' potential underutilized.
\end{itemize}
Our implementation of GA, BO and NEAT follows \citep{bhatia2021evolution}, whereas the implementation of MorphVAE follows \citep{song2024morphvae}. RoboGAN was originally proposed for the design automation of rigid robots, which we re-implemented on soft robots. 

\begin{table*}[htbp]
    \centering
    \caption{\textbf{Results of the ablation study. }Best fitness achieved at the end of evolution is compared between both of the two variants and their ablated versions. }
    \label{tab:ablation}
    \begin{tabular}{@{}c@{}}  
    \includegraphics[width=0.8\textwidth]{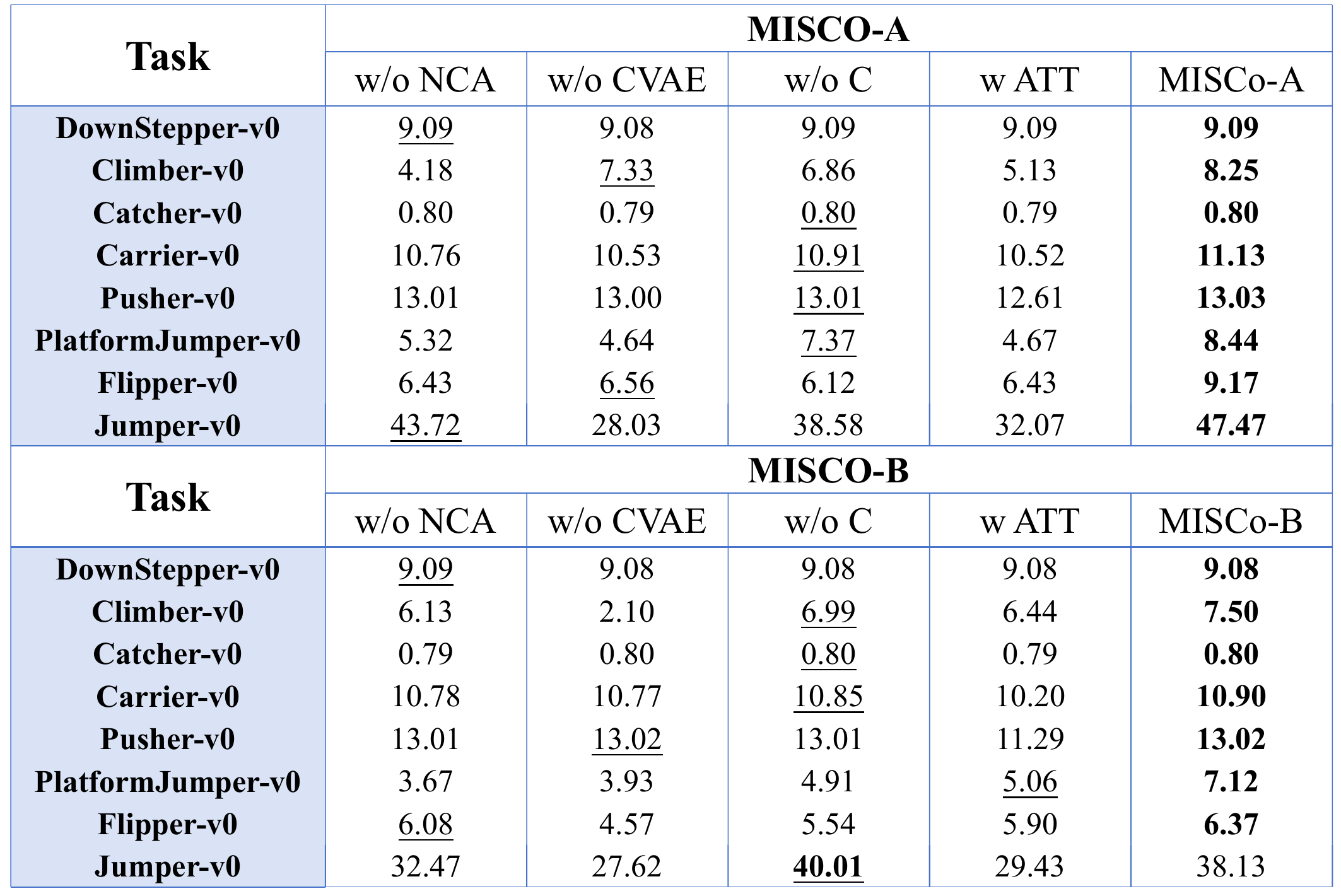}  
    \end{tabular}
\end{table*}

\section*{\textbf{Supplementary Material F:} \\Ablation study}
\label{appendix:ablation}
In the ablation study, we systematically modify our model by removing or replacing key components one at a time to evaluate their impact on optimization performance. Below is an overview of all modifications made in ablation study. 

\begin{itemize}
    \item \textbf{MISCO-w/o-VAE}: We replace our VAE with a simple statistical model -- a multinomial distribution that assumes independence across different voxel positions (\emph{i.e.,} univariate factorization). This would provide insights into how probabilistic generative models contribute to efficient optimization with their substantially stronger representational capacity. 
    \item \textbf{MISCO-w/o-NCA}: In this case, the inter-voxel coordination mechanism is entirely removed, which allows us to evaluate the importance of explicitly characterizing inter-dependencies across different voxels within a robot. 
    \item \textbf{MISCO-w-ATT}: Here NCA is replaced with a different message-passing scheme, \emph{i.e.,} the attention mechanism. Such a substitution is intended to test the robustness of our algorithmic design when one of the original components is replaced with an alternative that serves a similar purpose. 
    \item \textbf{MISCO-w/o-C}: In this case, task embedding is removed from the input of the approximate posterior in VAE, rendering the posterior distribution unconditional on task types. This ablated version thus serves to verify the necessity to condition not only the generative process, but also variational inference on tasks in a multi-task learning scheme. 
\end{itemize} 

Here we present the detailed results of ablations and on this basis, further make a brief comparative analysis between MISCO-A and MISCO-B in terms of evolutionary strategies.

The ablation results of MISCO-A and MISCO-B shown in Supplementary Table \ref{tab:ablation} reveal consistent patterns across almost all of eight task environments, highlighting the synergy and necessity of each component. When individual components are removed, performance degradation is observed, reaffirming that each element plays a critical role in maintaining stability and adaptability. A notable exception arises in the "Flipper-v0" task, where an ablated version of MISCO-B surpasses the complete model. However, even in this scenario, the ablated version falls short of the performance achieved by MISCO-A. 

The broader trends across the remaining tasks suggest that while MISCO-B is competent, it lacks the same evolutionary efficiency found in MISCO-A. The synergy effect of components in MISCO-B is effective, but the absence of MISCO-A’s focus on elite utilization strategy limits its ability to reach optimal solutions within shorter timeframes. This difference not only affects the quality of the search but also hinders MISCO-B’s ability to maintain consistent performance across challenging environments.

By contrast, one distinguishing strength of MISCO-B lies in its capacity for exploration and diversity improvement. In our design, randomness is intentionally introduced into the natural selection process during evolution, and the degree of randomness is governed by a tunable hyperparameter. This level of flexibility ensures that the search process can dynamically adjust between prioritizing elite solutions and expanding the diversity of potential candidates, empowering researchers to tailor the exploration-exploitation trade-off according to the specific demands of the task

The comparison of these two variants underscores the importance of tailoring evolutionary strategies to the nature of the problem: when efficiency is critical, MISCO-A excels; when diversity is more valuable, MISCO-B provides an effective alternative.

\section*{\textbf{Supplementary Material G:} \\Implementation details}
\label{appendix:implementation}
Regarding the training of MEC-VAE, the number of robot samples selected in each model update is set to 50. The number of model updates in each generation is linearly increased from 50 to 250 to prevent premature convergence. The number of inter-voxel coordination steps $K$ in NCA is set to 30. We employ the Proximal Policy Optimization (PPO; \citealp{schulman2017proximal}) algorithm -- a prominent reinforcement learning algorithm -- for controller optimization and robot fitness evaluation. All multi-layer perceptrons (MLPs) involved in this work, including the actor and critic networks in PPO and those in MEC-VAE, consist of two hidden layers, each containing 64 units and using the Tanh activation function. The experiments are conducted on a server equipped with Intel Xeon processors running at 2.20 GHz and four NVIDIA Tesla RTX GPUs, operating under Ubuntu 22.04. Our code is available at \url{https://github.com/xh621/MISCO} for reproducibility. Detailed parameter settings are listed in Supplementary Table \ref{table:problem_definition}, \ref{table:model_parameters} and \ref{table:mlp_architecture}. 

\begin{table}[htbp]
\centering
\caption{\textbf{Definition of VSR design problem.}}
\begin{tabular}{@{}c c@{}}
\toprule
\textbf{Hyperparameter} & \textbf{Value} \\ \midrule
Number of tasks optimized together in multi-task learning  & 3              \\
Size of robot design space    & 5$\times$5     \\
Number of voxel type            & 5              \\ \bottomrule
\end{tabular}
\label{table:problem_definition}
\end{table}

\begin{table}[htbp]
\centering
\caption{\textbf{Model hyperparameters.}}
\begin{tabular}{@{}c c c@{}}
\toprule
\textbf{Hyperparameter}          & \textbf{Notation} & \textbf{Value} \\ \midrule
Task embedding dimension         & $Y_t$                & 128                \\
Task latent dimension            & $H_t$                & 128                \\
Position embedding dimension     & $P_v$                & 128                \\
Voxel-specific latent dimension  & $h_v$                & 64                 \\
Voxel-specific steady state dimension  & $H_v$            & 64                 \\ \bottomrule
\end{tabular}
\label{table:model_parameters}
\end{table}

\begin{table}[htbp]
\centering
\caption{\textbf{MLP architectures.}}
\captionsetup{skip=0pt}  
\begin{tabular}{@{} c >{\centering\arraybackslash}p{0.6\columnwidth} @{}} 
\toprule
\textbf{MLP Type}                & \textbf{Structure}           \\ \midrule
MLP in task awareness            & {[}input size, 128, ReLU, 128, ReLU, 128, ReLU, output size, output transform{]} \\
MLP in NCA                       & {[}input size, 128, ReLU, 128, ReLU, 128, ReLU, output size{]}   \\
MLP in shared output layer       & {[}input size, 128, ReLU, 128, ReLU, 128, ReLU, output size{]}   \\
MLP in encoder                   & {[}input size, 128, ReLU, 128, ReLU, 128, ReLU, output size, output transform{]} \\ \bottomrule
\end{tabular}
\label{table:mlp_architecture}
\end{table}

\textit{Note:} For standard MLPs, the output activation is set to \textbf{None}. However, for MLPs used to generate variance $\Sigma$ of task latent distribution, an \texttt{Exp} function is applied to the output layer to ensure non-negativity.

\bibliographystyle{IEEEtran}
\bibliography{sn-bibliography}
\end{document}